\documentclass[lettersize,journal]{IEEEtran}
\usepackage{amsmath,amsfonts}
\usepackage{algorithmic}
\usepackage{algorithm}
\usepackage{array}
\usepackage[caption=false,font=normalsize,labelfont=sf,textfont=sf]{subfig}
\usepackage{textcomp}
\usepackage{stfloats}
\usepackage{url}
\usepackage{verbatim}
\usepackage{graphicx}
\usepackage{cite}
\usepackage[table]{xcolor}
\usepackage{booktabs}
\usepackage{multirow}
\usepackage{siunitx}
\usepackage{adjustbox}
\usepackage{caption}
\usepackage{amssymb}
\usepackage{diagbox}
\usepackage{makecell}
\usepackage{pifont}
\usepackage[flushleft]{threeparttable}

\usepackage[compact]{titlesec}
\titlespacing{\section}{0pt}{*0.8}{*0.8}
\titlespacing{\subsection}{0pt}{*0.6}{*0.6}
\titlespacing{\subsubsection}{0pt}{*0.7}{*0.7}

\def\IEEEtitletopspaceextra{-12pt}

\newif\ifrevblue \revbluefalse
\newcommand{\rev}[1]{\ifrevblue\textcolor{blue}{#1}\else#1\fi}

\begin{document}
\bstctlcite{BSTcontrol}

\title{BEV-ODOM2: Enhanced BEV-based Monocular Visual Odometry with PV-BEV Fusion and Dense Flow Supervision for Ground Robots}

\author{Yufei Wei$^{1}$, Chenxiao Hu$^{1}$, Wangtao Lu$^{1}$, Sha Lu$^{1}$, Yuxiang Cui$^{2}$, Fuzhang Han$^{2}$, Rong Xiong$^{1}$ and Yue Wang$^{1\dagger}$\vspace{-16pt}%
\thanks{$^{1}$Laboratory of Industrial Control and Technology, and the Institute of Cyber-Systems and Control, Zhejiang University, Hangzhou, 310058, China.}%
\thanks{$^{2}$Zhejiang Humanoid Robot Innovation Center Co., Ltd., Ningbo, Zhejiang 315012, China.}%
\thanks{$^{\dagger}$Corresponding author wangyue@iipc.zju.edu.cn.}%
}

\markboth{}%
{Wei \MakeLowercase{\textit{et al.}}: BEV-ODOM2: Enhanced BEV-based Monocular Visual Odometry}

\maketitle

\begin{abstract}
Scale-consistent ego-motion estimation is fundamental for autonomous ground robots. Bird's-Eye-View (BEV) representation naturally addresses the scale drift problem of monocular visual odometry (MVO) by providing a metric-scaled planar workspace, enabling the simplification of 6-DoF ego-motion to a more robust 3-DoF model. However, existing BEV-based methods suffer from two key limitations: sparse supervision signals from pose-only training, and information loss during perspective-to-BEV projection. We present BEV-ODOM2, an enhanced framework in the BEV-ODOM line that addresses both limitations without supervision modalities beyond the pose ground truth. Our approach introduces (1) pose-derived dense rigid BEV flow supervision, which reparameterizes the 3-DoF pose ground truth into a pixel-level training signal, and (2) Perspective View (PV)-BEV fusion, which computes correlation volumes before projection \rev{to retain additional motion cues and help alleviate projection ambiguity}. An enhanced rotation sampling strategy further balances diverse motion patterns during training. We evaluate on four datasets with varied spatial scales: KITTI, Oxford, NCLT, and our newly collected ZJH-VO benchmark. \rev{BEV-ODOM2 reduces the RTE of BEV-ODOM by 39\% on average across the four datasets.} It further enables closed-loop navigation on a physical robot with centimeter-level cross-track error. Real-time inference on an NVIDIA Jetson AGX Orin confirms edge deployment feasibility. The code and the ZJH-VO dataset are publicly released to facilitate future research.
\end{abstract}

\vspace{-0.15cm}

{\small\noindent\textbf{\textit{Note to Practitioners}}---This paper shows how a ground robot can track its own position accurately with a single ordinary camera, using a method designed to be cheap to deploy. Unlike methods that need a laser scanner, a depth camera, or hand-labeled training data, ours is trained only from the position logs a robot already records, so adopting it adds little data-collection or labeling work. It runs in real time at over 20 frames per second \rev{with 224$\times$384 input images and an accelerated backbone} on a small embedded computer with modest memory\rev{. It can therefore} be added to existing indoor service, inspection, or outdoor platforms without new hardware. A practical near-term use is \rev{as a backup localization source when GPS drops out indoors, in tunnels, or in garages. In this role, the method keeps the sideways drift of the robot within typical path tolerances for about ten seconds}. The main limitation is that the robot should move on a mostly flat surface, since steep slopes or strong bumps lower its accuracy. We release the code and the dataset for easy reuse.\par}


\section{Introduction}

\IEEEPARstart{R}{eliable} ego-motion estimation is a fundamental capability for autonomous ground robots across a wide range of automation tasks, from outdoor mobile platforms and field robotics to indoor service and inspection robots operating in structured environments. Monocular Visual Odometry (MVO) \cite{scaramuzza2011visual, chen2020survey} provides a lightweight solution for this task, yet conventional methods suffer from cumulative scale drift that degrades localization over long trajectories. Bird's-Eye-View (BEV) representation has emerged as a powerful paradigm to mitigate this challenge \cite{ross2022bev, zhang2025bev}. For ground robots, motion is predominantly planar \cite{unger2024multi}, allowing pose estimation to be simplified from six degrees of freedom (6-DoF) to a more robust 3-DoF model (x, y, yaw). This simplification naturally aligns with the unified, metric-scaled grid of BEV representation \cite{wei2025bev}, reducing computational complexity and mitigating error accumulation in non-primary motion axes.

\begin{figure}[t]  %
\centering
\includegraphics[width=0.46\textwidth]{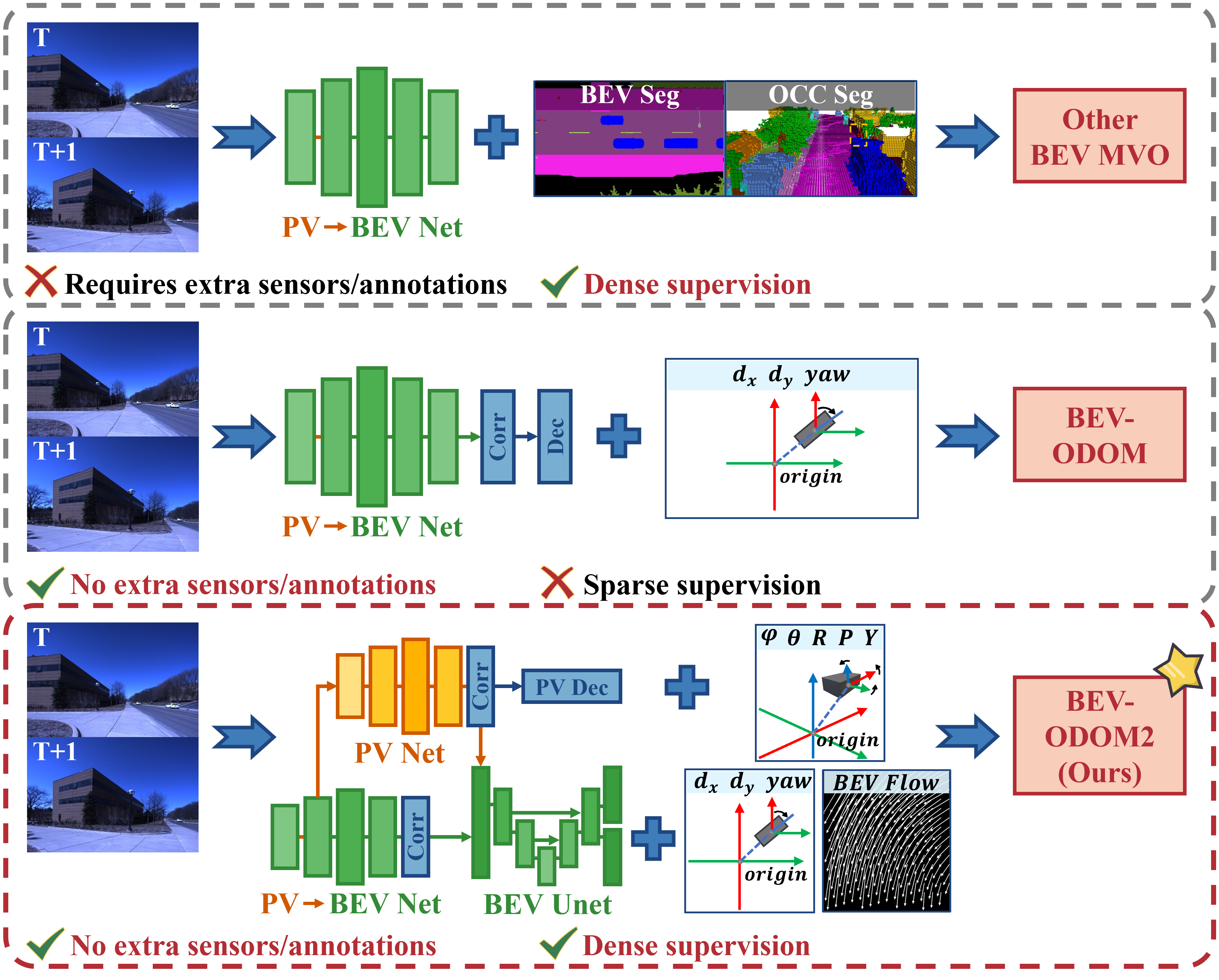}
\caption{Comparison of BEV-based monocular odometry methods. Unlike prior approaches needing extra annotations or limited by sparse supervision, our BEV-ODOM2 leverages PV-BEV fusion and dense rigid BEV flow to provide rich supervision from pose data alone.}
\label{fig:fig1}
\vspace{-0.6cm}
\end{figure}

BEV-ODOM \cite{wei2024bev} demonstrates that a BEV-based framework can effectively reduce scale drift using only 3-DoF pose supervision, eliminating the need for auxiliary tasks like depth estimation \cite{tateno2017cnn} or semantic segmentation \cite{bescos2018dynaslam}, which add complexity and data collection costs. However, this and other BEV-based approaches face two critical limitations. First, the supervision signal from a single 3-DoF pose vector is sparse, providing insufficient guidance for the network to learn fine-grained, pixel-level correspondences between consecutive BEV feature maps. As recent works like DUSt3R \cite{wang2024dust3r} have shown, dense supervision is crucial for achieving a robust geometric understanding of motion. Second, the standard transformation from PV to BEV, often implemented with architectures like Lift-Splat-Shoot (LSS) \cite{philion2020lift, li2023bevdepth}, inherently causes information loss \rev{and projection ambiguity. The projection places image features by the predicted depth and a local ground-plane prior. When bumps or the kinematic response of the vehicle body violate this prior, the BEV correlation no longer maps uniquely to the planar motion. The network then learns inconsistent motion patterns, which degrades performance.} Fig.~\ref{fig:fig1} contrasts our approach with prior BEV-based methods.

To overcome these challenges, \rev{we present BEV-ODOM2, a unified end-to-end monocular visual odometry framework built on two strategies that introduce no new types of external supervision.} First, to address the sparse supervision problem, we introduce pose-derived dense rigid BEV flow supervision. Because the BEV grid is metric-scaled, a 3-DoF relative pose defines a per-pixel displacement field over the grid, which provides a dense training signal that guides the network to learn detailed feature correspondences. Second, we introduce a PV-BEV dual-branch fusion strategy to counter \rev{the} projection ambiguity. \rev{A parallel branch computes a correlation-based cost volume from the PV features \cite{sun2018pwc, teed2020raft} before the LSS projection, so its matching scores depend on neither the predicted depth nor the ground-plane prior. The branch then projects this volume into BEV space and fuses it with the BEV-native correlation. The fusion thus retains additional motion cues and helps alleviate the projection ambiguity. Together, the two strategies improve accuracy while preserving the scale consistency and drift suppression inherent to BEV-based odometry. An enhanced rotation sampling strategy further balances diverse motion patterns during training.}

We also introduce ZJH-VO, a new ground-robot odometry dataset covering multiple scenes and scales, addressing the lack of publicly available benchmarks with such diversity. We use it to evaluate BEV-ODOM2 \rev{on unseen trajectories from four scene types} and release it to support future research.

Our contributions can be summarized as follows:
\begin{itemize}
\item We present BEV-ODOM2, an extension of the BEV-ODOM line of BEV-based MVO. Using pose supervision alone, it delivers strong overall accuracy while preserving scale consistency.

\item We introduce two components that address the two limitations of prior BEV-based methods: a pose-derived dense rigid BEV flow supervision, which turns the pose ground truth into dense pixel-level guidance for correspondence learning, and a PV-BEV dual-branch fusion, which computes feature-level correlation before the LSS projection to retain \rev{additional motion cues and help alleviate projection ambiguity}.

\item We develop data augmentation techniques that target biases common in ground-robot datasets, including enhanced sampling of large rotations and extended temporal sampling for robustness to diverse motion dynamics.

\item We construct the ZJH-VO dataset, a multi-scene multi-scale benchmark for ground-robot odometry covering both outdoor and indoor scenarios with diverse motion characteristics. We evaluate BEV-ODOM2 both offline on four datasets (KITTI \cite{geiger2013vision}, Oxford \cite{barnes2020oxford}, NCLT \cite{carlevaris2016university}, ZJH-VO) and online in closed-loop navigation on a physical robot\rev{. BEV-ODOM2 improves over BEV-ODOM in both accuracy and task-level reliability}. Edge deployment on an NVIDIA Jetson AGX Orin further confirms real-time feasibility on resource-constrained embedded platforms. The source code and the ZJH-VO dataset are available at \url{https://github.com/WeiYuFei0217/BEV-ODOM2} and \url{https://github.com/WeiYuFei0217/ZJH-VO-Dataset}.

\end{itemize}

\section{Related Work}
This section reviews existing literature on visual odometry, categorized into traditional geometric methods, learning-based methods in the perspective view, and emerging learning-based methods that leverage the BEV representation.

\subsection{Traditional Methods}
Traditional MVO methods are founded on geometric principles and are highly interpretable. These approaches are typically classified as feature-based, direct, or semi-direct. Feature-based methods, such as ORB-SLAM3 \cite{campos2021orb}, detect and match sparse keypoints across consecutive frames to estimate ego-motion. Direct methods, like DSO \cite{engel2017direct}, minimize photometric errors over a set of pixels, offering high precision in textured environments. Semi-direct methods, exemplified by SVO \cite{forster2014svo}, combine the efficiency of direct tracking with the robustness of feature-based optimization. Recent advances improve robustness by replacing handcrafted features with learned descriptors and adaptive motion models \cite{han2024basl}, yet their fundamental dependence on initial depth estimates for scale establishment remains.

These initial depth estimates, taken from the first few frames, serve as global scale references. Calibration errors, feature mismatches, and motion blur then accumulate during iterative pose updates, resulting in severe scale drift that limits their effectiveness in long-distance navigation tasks.

\subsection{Learning-Based Methods Under Perspective View}
Learning-based PV methods have emerged as a powerful alternative, leveraging deep neural networks to overcome the limitations of traditional approaches, particularly in scale estimation. Early works, such as DeepVO \cite{wang2017deepvo}, demonstrate the feasibility of using end-to-end recurrent convolutional neural networks to directly regress pose from image sequences. More recently, VoT \cite{yugay2025visual} follows the same paradigm and regresses relative pose with a transformer over pretrained visual features. However, direct regression offers limited geometric interpretability, which can constrain generalization.

Another line of research integrates geometric constraints with deep learning to improve upon direct pose regression. One prominent approach leverages depth and optical flow to provide the network with explicit geometric cues. For instance, TartanVO \cite{wang2021tartanvo} and DF-VO \cite{zhan2021df} use them as auxiliary supervision for scale anchoring and accuracy enhancement, and VOLDOR \cite{min2020voldor} infers pose directly from externally estimated dense optical flow.
A further development in this area involves tightly coupling learned feature representations with geometric optimization within a single architecture. DROID-SLAM \cite{teed2021droid} combines dense correlation volumes with a differentiable bundle adjustment layer, and its successor DPVO \cite{teed2024deep} applies this design to sparse image patches, enabling the joint optimization of pose and structure. SSVO \cite{chen2024salient} extends this design to pose-only supervision through self-supervised homographic pretraining.

A more recent line builds on large pretrained 3D reconstruction models. MASt3R-SLAM \cite{murai2025mast3r} and VGGT-SLAM \cite{maggio2026vggt} run zero-shot from such priors and recover strong geometry, at the cost of heavy computation. \rev{GRS-SLAM3R \cite{shen2025grs} uses a gated recurrent spatial memory and submap alignment for global consistency.}

Despite their significant advancements, a common limitation of these perspective-view methods is their reliance on auxiliary signals, such as depth, optical flow, or pretrained 3D priors, to ensure scale consistency. This reliance increases the cost and complexity of data collection and annotation. Methods trained with pose supervision alone avoid this cost, but the perspective view provides no metric reference, so their scale drifts or must be recovered by alignment with ground truth.

\begin{figure*}[t]
\centering
\includegraphics[width=1.0\textwidth]{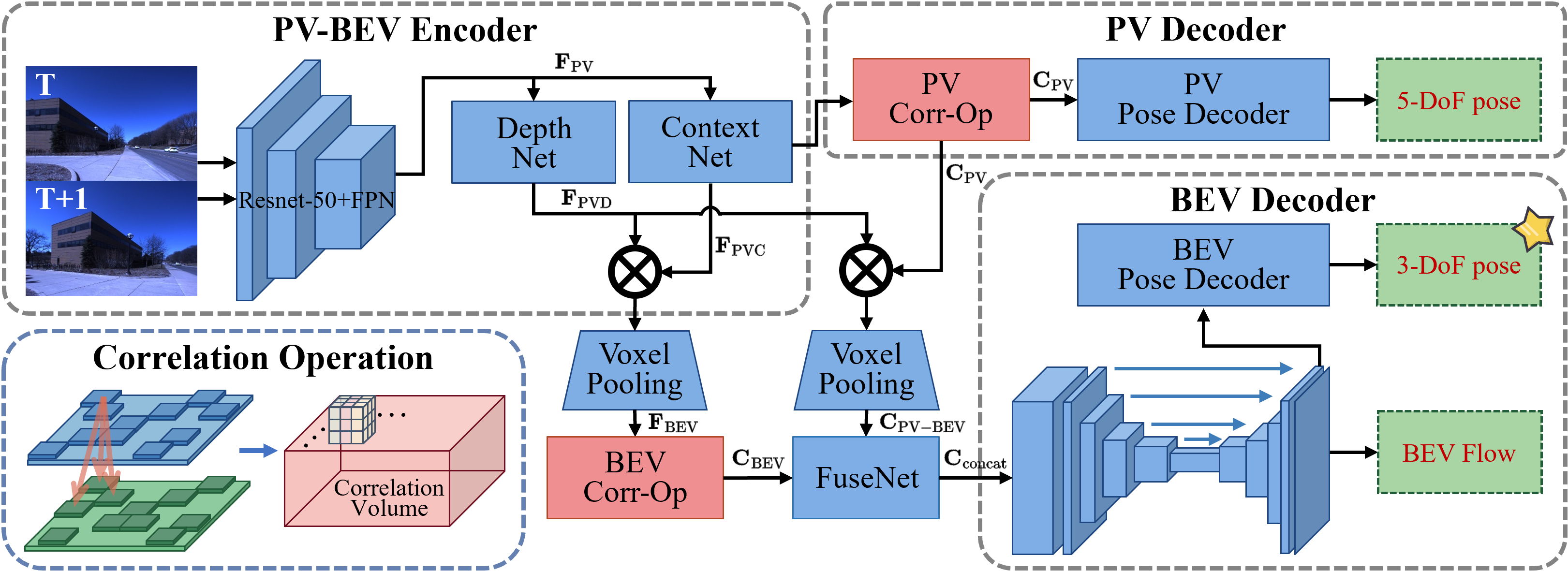}
\caption{Overview of the BEV-ODOM2 framework. The \textbf{PV-BEV Encoder} shares a ResNet-50+FPN backbone and projects both PV features and PV correlation volumes into BEV space via LSS. The \textbf{PV Decoder} regresses a 5-DoF pose from PV correlations. The \textbf{BEV Decoder} fuses projected PV and BEV-native correlations through a FuseNet to jointly predict the dense rigid BEV flow and the final 3-DoF pose.}
\label{fig:framework}
\vspace{-0.3cm}
\end{figure*}

\subsection{Learning-Based Methods Under BEV Representation}
\rev{Robotics uses a range of full 3D scene representations, from point clouds and neural radiance fields to the pretrained reconstruction models above \cite{deng2025best}. BEV is more compact: it keeps only the metric planar layout of the scene and drops the explicit height axis.}
Leveraging the common ground-plane assumption for ground robots, BEV representation has emerged as a promising paradigm for MVO \cite{wei2024bev, wei2025bev}.
The inherent properties of BEV's unified metric-scaled grid enable a natural reduction from 6-DoF to 3-DoF pose estimation, establishing an implicit scale anchoring mechanism through the consistent spatial resolution of the grid structure.

Initial BEV-based odometry approaches primarily focus on semantic-level understanding for motion estimation.
BEV-SLAM \cite{ross2022bev} constructs semantic BEV representations through CNNs and performs localization by matching against pre-existing semantic maps.
BEV-Locator \cite{zhang2025bev} transforms perspective imagery into BEV encodings and employs cross-modal transformers for semantic feature alignment.
OCC-VO \cite{li2023occ} extends this paradigm to volumetric space, generating 3D semantic occupancy representations from multi-view inputs for global map registration.

While these approaches demonstrate the potential of BEV, their performance is fundamentally tied to the accuracy of the intermediate map generation, which requires costly and extensive annotation for supervision.

Recognizing these limitations, recent advances explore direct motion estimation in BEV space without semantic dependencies. BEV-ODOM \cite{wei2024bev} demonstrates that a BEV-based framework using correlation-based regression can effectively learn to estimate ego-motion with strong scale consistency using pose supervision alone. Subsequently, BEV-DWPVO \cite{wei2025bev} improves upon this by replacing the regression backend with a more interpretable pipeline, which extracts and matches keypoints in the BEV space and computes the final pose using a differentiable weighted Procrustes solver \cite{gower1975generalized}.

While these pose-only methods achieve notable scale consistency, they directly optimize with sparse pose supervision without leveraging the constructible nature of BEV representation for richer guidance.
Additionally, \rev{bumps and the kinematic response of the vehicle body during operation violate the local ground-plane prior of the perspective-to-BEV transformation}, potentially degrading accuracy in complex driving scenarios.
These limitations motivate our proposed BEV-ODOM2, which constructs dense supervision from pose ground truth and employs a dual-branch architecture to \rev{retain} perspective-view motion cues while maintaining the scale consistency benefits of BEV representation.

\section{Methodology}
\label{sec:method}
Ground robots in autonomous automation tasks primarily operate under constrained planar motion, enabling the simplification of pose estimation from 6-DoF to 3-DoF while focusing on the essential motion components for ego-motion estimation \cite{wei2024bev, unger2024multi}. Building upon this principle, BEV-ODOM2 addresses two critical limitations in existing approaches: sparse supervision signals that inadequately guide correlation learning \cite{chen2020survey}, and \rev{the information loss and projection ambiguity of the} perspective-to-BEV transformation. Our solution combines dense rigid BEV flow supervision with PV-BEV dual-branch fusion, where all supervision signals are constructed directly from the pose ground truth.

\subsection{Overall Framework}
\label{subsec:framework}
Fig.~\ref{fig:framework} illustrates the overall architecture of BEV-ODOM2. The system processes consecutive monocular images $I_t$ and $I_{t+1}$ through parallel processing branches that extract complementary motion information. The PV branch computes correlation before the LSS transformation \cite{philion2020lift}, while the BEV branch performs correlation on unified metric-scaled features for final 3-DoF pose estimation. The key design lies in projecting PV-derived correlation cost volumes into BEV space through the LSS pipeline, and then concatenating them with BEV correlation cost volumes to fuse relative motion features from both representations.

The framework employs three \rev{supervision signals}, all derived from pose ground truth: (1) dense rigid BEV flow supervision that exploits the constructible property of the BEV representation for pixel-level motion learning, (2) 5-DoF pose supervision for the PV branch (excluding scale to maintain consistency with monocular constraints), and (3) 3-DoF pose supervision for the final output. \rev{This scheme requires no supervisory modality beyond the pose ground truth.}

\subsection{Feature Extraction and PV Branch}
\label{subsec:pv_processing}
The dual-branch architecture leverages complementary advantages of perspective view and BEV representations to address \rev{the information loss and projection ambiguity of BEV-only approaches}. Both branches share a common feature extraction backbone to ensure consistent feature representation while processing motion information through distinct geometric paradigms.

\textbf{Shared Feature Extraction:} The system employs a ResNet-50 \cite{he2016deep} backbone integrated with Feature Pyramid Network (FPN) \cite{lin2017feature} to extract multi-scale hierarchical features $\mathbf{F}_{\text{IF}} \in \mathbb{R}^{C_{\text{IF}} \times H_{\text{PV}} \times W_{\text{PV}}}$ from consecutive input images $I_t$ and $I_{t+1}$. Camera intrinsic parameters $\mathbf{K} \in \mathbb{R}^{3 \times 3}$ and extrinsic parameters $\mathbf{E} \in \mathbb{R}^{3 \times 4}$ are encoded through a Multi-Layer Perceptron and fused with image features via element-wise multiplication:

\begin{equation}
\mathbf{F}_{\text{PV}} = \mathbf{F}_{\text{IF}} \odot \text{MLP}(\mathbf{K}, \mathbf{E}),
\end{equation}
where $\odot$ denotes element-wise multiplication and $\mathbf{F}_{\text{PV}} \in \mathbb{R}^{C_{\text{PV}} \times H_{\text{PV}} \times W_{\text{PV}}}$ represents the processed perspective features.

\textbf{Perspective View Branch Processing:} \rev{The PV branch computes a correlation volume before the LSS transformation and is supervised by a 5-DoF pose objective. The BEV correlation observes motion only after the projection, which places features by the predicted depth and a local ground-plane prior. During operation, bumps and the kinematic response of the vehicle body make this prior hold only approximately. The same scene content can then fall into shifted BEV cells, and the BEV correlation no longer maps uniquely to the planar motion. The PV correlation is computed on the image grid, so its matching scores rely on neither the predicted depth nor the ground-plane prior. \rev{It thus helps compensate for this cell misplacement in the BEV correlation. It also perceives rotation more directly, because a camera rotation shifts the whole image regardless of scene depth, whereas translation moves image points by amounts that depend on depth.} Fusing the two correlations therefore retains additional motion cues and helps alleviate this projection ambiguity.} Following established precedents in optical flow estimation \cite{sun2018pwc, teed2020raft}, local correlation computation between consecutive perspective features captures pixel-level motion patterns:

\begin{equation}
\mathbf{C}_{\text{PV}}[{\scriptstyle \Delta x, \Delta y, x, y}] = \sum_{c=1}^{C_{\text{PV}}} \mathbf{F}_{\text{PV}}^{t}[{\scriptstyle c, x, y}] \cdot \mathbf{F}_{\text{PV}}^{t+1}[{\scriptstyle c, x + \Delta x, y + \Delta y}],
\end{equation}
where $(x, y)$ is a pixel of the PV feature map. The offsets $\Delta x, \Delta y \in [-\Delta_{\text{PV}}, \Delta_{\text{PV}}]$ define the local search range with $\Delta_{\text{PV}} = 5$, producing correlation volume $\mathbf{C}_{\text{PV}} \in \mathbb{R}^{(2\Delta_{\text{PV}}+1)^2 \times H_{\text{PV}} \times W_{\text{PV}}}$. \rev{Each channel of $\mathbf{C}_{\text{PV}}$ corresponds to one image-plane offset $(\Delta x, \Delta y)$, and each spatial location corresponds to an anchor pixel of frame $t$.}

\textbf{5-DoF PV Pose Supervision:} \rev{To provide an additional pose signal for the PV branch,} the PV correlation volume undergoes spatial dimension reduction through a sequence of convolutional layers followed by fully connected layers for pose regression:
\begin{equation}
\mathbf{T}_{\text{5-DoF}} = \text{MLP}_\text{PV}(\text{CNN}_\text{PV}(\mathbf{C}_{\text{PV}})),
\end{equation}
where $\text{CNN}_\text{PV}$ represents sequential convolutional downsampling and $\text{MLP}_\text{PV}$ indicates fully connected layers for pose estimation. \rev{The output $\mathbf{T}_{\text{5-DoF}} = [\mathbf{R}_{\text{pred}}, \mathbf{t}_{\text{pred}}]$ consists of a rotation matrix $\mathbf{R}_{\text{pred}} \in SO(3)$ and a translation vector $\mathbf{t}_{\text{pred}} \in \mathbb{R}^{3}$. The MLP outputs four rotation parameters and the three-dimensional $\mathbf{t}_{\text{pred}}$. The rotation parameters are normalized to a unit quaternion and converted to $\mathbf{R}_{\text{pred}}$. The loss in Section~\ref{subsec:loss_function} uses only the normalized translation direction $\hat{\mathbf{t}}_{\text{pred}}$, so scale is excluded to maintain consistency with monocular visual constraints. The three rotational and two directional degrees of freedom form the 5-DoF target.}

\subsection{BEV Projection Pipeline}
\label{subsec:bev_projection}
The BEV projection process transforms both perspective view features and correlation volumes into a unified metric-scaled representation while preserving essential motion information from different geometric paradigms. This transformation employs the LSS architecture \cite{philion2020lift} with shared network parameters to ensure consistent geometric mapping across different input modalities.

\textbf{Perspective Feature to BEV Transformation:} The processed perspective features $\mathbf{F}_{\text{PV}} \in \mathbb{R}^{C_{\text{PV}} \times H_{\text{PV}} \times W_{\text{PV}}}$ undergo depth-aware projection to generate BEV representation. Two specialized networks further refine the perspective features to predict contextual features and depth distributions, outputting $\mathbf{F}_{\text{PVC}} \in \mathbb{R}^{C_{\text{PV}} \times H_{\text{PV}} \times W_{\text{PV}}}$ and $\mathbf{F}_{\text{PVD}} \in \mathbb{R}^{D_{\text{PV}} \times H_{\text{PV}} \times W_{\text{PV}}}$ respectively, where $D_{\text{PV}}$ represents the discretized depth resolution. The contextual and depth features are dimension-expanded to enable element-wise multiplication across channel and depth dimensions:
\begin{equation}
\mathbf{F}_{\text{PVmulti}} = \mathbf{F}'_{\text{PVC}} \odot \mathbf{F}'_{\text{PVD}},
\end{equation}
where $\mathbf{F}_{\text{PVC}}$ and $\mathbf{F}_{\text{PVD}}$ are expanded to $C_{\text{PV}} \times 1 \times H_{\text{PV}} \times W_{\text{PV}}$ and $1 \times D_{\text{PV}} \times H_{\text{PV}} \times W_{\text{PV}}$, and are referred to as $\mathbf{F}'_{\text{PVC}}$ and $\mathbf{F}'_{\text{PVD}}$ respectively, and $\odot$ denotes element-wise multiplication. This operation produces $\mathbf{F}_{\text{PVmulti}} \in \mathbb{R}^{C_{\text{PV}} \times D_{\text{PV}} \times H_{\text{PV}} \times W_{\text{PV}}}$, which combines contextual features across channels and depth for each pixel. Subsequently, frustum projection maps features at different depths to 3D voxel space, transforms them to the vehicle coordinate system using camera intrinsics and extrinsics, and accumulates them onto the 2D BEV plane. Voxel pooling aggregates these projected features, forming the final BEV feature representation $\mathbf{F}_{\text{BEV}} \in \mathbb{R}^{C_{\text{BEV}} \times H_{\text{BEV}} \times W_{\text{BEV}}}$, where $C_{\text{BEV}} = C_{\text{PV}}$ due to the feature channel preservation during the projection process.

\textbf{PV Correlation Volume to BEV Projection:} A central element of our design lies in projecting the PV correlation volume $\mathbf{C}_{\text{PV}} \in \mathbb{R}^{(2\Delta_{\text{PV}}+1)^2 \times H_{\text{PV}} \times W_{\text{PV}}}$ to BEV space using the identical LSS pipeline and network parameters. \rev{The projection process treats each correlation channel as a feature channel.} Since the correlation operation captures the motion of frame $t{+}1$ relative to the anchor frame $t$, the resulting volume is spatially anchored at frame $t$'s pixel grid and is therefore projected using frame $t$'s depth distribution:
\begin{equation}
\begin{aligned}
\mathbf{C}_{\text{PV-BEV}} &= \text{Project}_{\text{LSS}}(\mathbf{C}_{\text{PV}}, \mathbf{F}_{\text{PVD}}^{t}, \mathbf{K}, \mathbf{E}), \\
\rev{\mathbf{C}_{\text{PV-BEV}}[{\scriptstyle \delta, b}]} &\rev{= \sum_{p} \sum_{d} \mathbf{1}\big[\pi(p, d) \in b\big]\, \mathbf{F}_{\text{PVD}}^{t}[{\scriptstyle d, p}]\, \mathbf{C}_{\text{PV}}[{\scriptstyle \delta, p}],}
\end{aligned}
\end{equation}
where $\mathbf{K}$ and $\mathbf{E}$ represent camera intrinsic and extrinsic parameters, respectively. \rev{In the second line, $b$ is a BEV cell and $\delta = (\Delta x, \Delta y)$ is a correlation offset. The sum runs over anchor pixels $p$ and depth bins $d$, and $\pi(p, d)$ is the BEV cell of pixel $p$ lifted to depth $d$. The projection thus spreads the correlation vector of each anchor pixel over the BEV cells along its ray, weighted by the depth probability. Each BEV cell therefore holds the weighted PV matching patterns of the anchor pixels whose rays pass through it, not a 3D displacement. The depth of the matched endpoint is not needed. The BEV branch places the image feature of the same anchor with the same depth distribution. Each cell of $\mathbf{C}_{\text{PV-BEV}}$ is thus aligned with the same cell of the BEV correlation, which enables the channel-wise fusion in Section~\ref{subsec:motion_extraction_fusion}.}

\subsection{Motion Information Extraction and Fusion}
\label{subsec:motion_extraction_fusion}
The motion information extraction and fusion process combines correlation-based motion patterns from both BEV and projected PV representations to achieve comprehensive motion understanding while maintaining the computational advantages of 3-DoF pose estimation.

\textbf{BEV Correlation Computation:} Local correlation computation between consecutive BEV features captures planar motion patterns directly in the unified metric-scaled coordinate system:
\begin{equation}
\mathbf{C}_{\text{BEV}}[{\scriptstyle \Delta u, \Delta v, u, v}] = \sum_{c=1}^{C_{\text{BEV}}} \mathbf{F}_{\text{BEV}}^{t}[{\scriptstyle c, u, v}] \cdot \mathbf{F}_{\text{BEV}}^{t+1}[{\scriptstyle c, u + \Delta u, v + \Delta v}],
\end{equation}
where $(u, v)$ is a BEV cell. The offsets $\Delta u, \Delta v \in [-\Delta_{\text{BEV}}, \Delta_{\text{BEV}}]$ define the local correlation search range with $\Delta_{\text{BEV}} = 5$, producing $\mathbf{C}_{\text{BEV}} \in \mathbb{R}^{(2\Delta_{\text{BEV}}+1)^2 \times H_{\text{BEV}} \times W_{\text{BEV}}}$. As in the PV branch, frame $t$ serves as the anchor frame and frame $t{+}1$ is shifted during correlation. This correlation volume directly encodes planar motion components essential for ground robot navigation.

\textbf{Multi-Modal Correlation Fusion:} The projected PV correlation volume $\mathbf{C}_{\text{PV-BEV}}$ and native BEV correlation volume $\mathbf{C}_{\text{BEV}}$ are concatenated along the channel dimension to create a comprehensive motion representation:
\begin{equation}
\mathbf{C}_{\text{concat}} = \text{Concat}(\mathbf{C}_{\text{BEV}}, \mathbf{C}_{\text{PV-BEV}}) \in \mathbb{R}^{C_{\text{total}} \times H_{\text{BEV}} \times W_{\text{BEV}}},
\end{equation}
where $C_{\text{total}} = (2\Delta_{\text{BEV}}+1)^2 + (2\Delta_{\text{PV}}+1)^2$. This concatenated representation combines the BEV correlation's scale-consistent planar motion with the \rev{additional motion cues retained by} the projected PV correlation.

\textbf{Dense Flow Prediction and 3-DoF Pose Estimation:} A UNet-style encoder-decoder architecture \cite{ronneberger2015u} processes the concatenated correlation features to simultaneously extract and fuse multi-modal motion information while predicting dense flow fields and final pose estimates. The encoder progressively captures global motion patterns from both BEV and projected PV correlations:
\begin{equation}
\mathbf{F}_{\text{Enc}} = \text{U}_\text{Enc}(\mathbf{C}_{\text{concat}}).
\end{equation}
The decoder then upsamples the encoded features back to the BEV resolution to predict the dense rigid BEV flow, and skip connections restore the spatial details lost during downsampling:
\begin{equation}
\mathbf{F}_{\text{flow}} = \text{U}_\text{Dec}(\mathbf{F}_{\text{Enc}}) \in \mathbb{R}^{2 \times H_{\text{BEV}} \times W_{\text{BEV}}},
\end{equation}
where $\mathbf{F}_{\text{flow}}$ represents the predicted dense rigid BEV flow field. Critically, the 3-DoF pose estimation branch connects to the penultimate layer of the flow decoder to leverage the enriched motion features:
\begin{equation}
\mathbf{T}_{\text{3-DoF}} = \text{MLP}_\text{BEV}(\text{CNN}_\text{BEV}(\mathbf{F}_{\text{Dec}}^{L-1})) = [\theta, t_x, t_y],
\end{equation}
where $\mathbf{F}_{\text{Dec}}^{L-1}$ denotes the second-to-last layer features of the $L$-layer decoder, and \rev{$\theta$, $t_x$, and $t_y$ estimate the yaw and planar translation of $\mathbf{T}_{\text{rel}}$ defined in Section~\ref{subsec:flow_supervision}}. The features at this layer fuse motion information from both correlation sources and are directly refined by the dense flow supervision. Regressing the pose from these features therefore benefits from both the multi-modal fusion and the pixel-level motion learning.

\begin{figure}[t]  %
\centering
\includegraphics[width=0.46\textwidth]{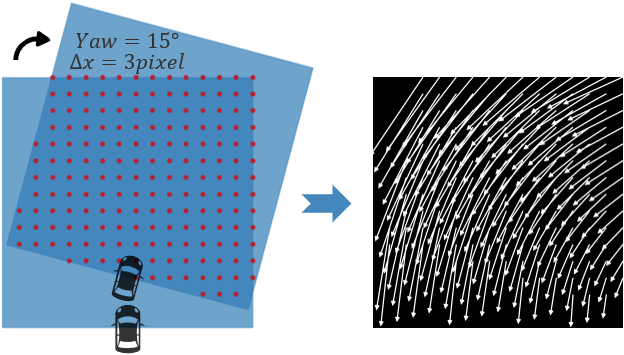}
\caption{Construction of the pose-derived dense rigid BEV flow ground truth. A 3-DoF relative pose (yaw and planar translation) defines a rigid transformation on the BEV grid (Eqs.~\ref{eq:vehicle_coordinates}--\ref{eq:optical_flow}), yielding a per-pixel displacement field (right) as dense supervision.}
\label{fig:flow}
\vspace{-0.6cm}
\end{figure}

\subsection{Pose-Derived Dense Rigid BEV Flow Supervision}
\label{subsec:flow_supervision}
Traditional pose supervision provides sparse signals that may be insufficient for learning the fine-grained motion patterns that benefit robust visual odometry \cite{chen2020survey}. We address this limitation with pose-derived dense rigid BEV flow supervision. All BEV cells lie in the ego vehicle coordinate system and share a consistent metric scale, so the 3-DoF pose ground truth defines a 2D rigid transformation of the entire BEV grid. Applying this transformation to every cell yields a dense displacement field, as illustrated in Fig.~\ref{fig:flow}. This field is a rigid flow implied by the pose, not an observed image-domain optical flow. \rev{It combines the relative pose label with the metric geometry of the BEV grid and requires no annotation beyond the pose ground truth.} Its construction therefore involves none of the depth and viewpoint reasoning that perspective-view optical flow requires, and it turns the existing pose ground truth into pixel-level motion guidance.

Given the relative pose transformation $\mathbf{T}_{\text{rel}} \in \mathbb{R}^{4 \times 4}$ between consecutive frames, the flow ground truth uses its planar part, namely the yaw rotation and planar translation. It is constructed in three steps: each BEV pixel is mapped to the vehicle coordinate system, transformed by $\mathbf{T}_{\text{rel}}$, and mapped back to pixel coordinates. \rev{We define $\mathbf{T}_{\text{rel}} = \mathbf{G}_{t+1}^{-1}\mathbf{G}_{t}$, where $\mathbf{G}_{t}$ is the vehicle pose at frame $t$. It maps points from the vehicle frame at $t$ to the vehicle frame at $t{+}1$. Its planar part is the target of the 3-DoF pose loss, and its full 3D form is the target of the 5-DoF pose loss.} For each BEV pixel at coordinates $(u, v)$, we first establish the corresponding homogeneous vehicle coordinates:

\begin{equation}
\mathbf{p}_{\text{veh}} = \begin{bmatrix} (o_y - v) \cdot r \\ (u - o_x) \cdot r \\ 0 \\ 1 \end{bmatrix},
\label{eq:vehicle_coordinates}
\end{equation}

where $r$ denotes the BEV resolution in meters per pixel, and $(o_x, o_y)$ represents the pixel coordinates of the vehicle coordinate system origin. \rev{In this mapping, $u$ increases to the right and $v$ increases downward. The $x$ axis thus points forward, i.e., up in the BEV image, and the $y$ axis points to the right in the image. The yaw $\theta$ of $\mathbf{T}_{\text{rel}}$ is positive when its rotation turns the $x$ axis toward the $y$ axis.} The rigid BEV flow is then computed as:

\begin{equation}
\mathbf{F}_{\text{flow-gt}}[u, v] = \mathcal{P}(\mathbf{T}_{\text{rel}} \mathbf{p}_{\text{veh}}) - [u, v],
\label{eq:optical_flow}
\end{equation}

where $\mathcal{P}(\cdot)$ denotes the projection operation that converts the transformed homogeneous vehicle coordinates back to pixel coordinates using the inverse mapping of Equation~\eqref{eq:vehicle_coordinates}. \rev{The target is thus defined on the grid of frame $t$ and gives the displacement of each cell into frame $t{+}1$ in grid units.}

Evaluating Equation~\eqref{eq:optical_flow} at every pixel yields the dense supervision target $\mathbf{F}_{\text{flow-gt}} \in \mathbb{R}^{2 \times H_{\text{BEV}} \times W_{\text{BEV}}}$, which requires no optical flow sensors or additional annotations. \rev{Every cell of the predicted field contributes to the loss. No visibility, dynamic-object, or boundary mask is applied, and a cell whose transformed position leaves the grid keeps its computed displacement.} Since every cell in this target follows one shared rigid motion, the network is trained to explain the scene with the dominant rigid ego-motion. Moving objects contradict this target only on the minority of cells they occupy, so their influence on training stays limited. We validate this dynamic-scene robustness in Section~\ref{subsubsec:dynamic}, and analyze the residual impact of the planar assumption on out-of-plane motion in Section~\ref{subsubsec:kitti_results}.

\subsection{Enhanced Rotation Sampling Strategy}
\label{subsec:enhanced_strategy}
To address dataset biases toward straight-line driving scenarios and improve robustness to diverse motion patterns in real-world deployment, we implement a targeted sampling strategy that ensures balanced representation of both linear and turning maneuvers while leveraging temporal consistency within driving sequences.

\textbf{Rotation-Aware Data Augmentation:} We construct comprehensive motion pattern databases by preprocessing training sequences to identify frames with diverse rotational characteristics. For each frame in the training dataset, we establish a temporal search window spanning one minute before and after the current timestamp, then extract relative pose transformations for all frame pairs within this window. Frame pairs are categorized based on two criteria: (1) relative yaw angle differences between 15° and 45° are stored in a high-rotation list $\mathcal{S}_{\text{high}}$, while pairs with angular differences below 15° are maintained in a standard-rotation list $\mathcal{S}_{\text{standard}}$, and (2) spatial displacement constraints where only pairs with translation distances up to 4 meters are retained to ensure meaningful correspondence learning.

During training, when a particular frame is selected, the system probabilistically samples from these pre-constructed lists using a 70:30 distribution, where 70\% of samples are drawn from $\mathcal{S}_{\text{high}}$ and 30\% from $\mathcal{S}_{\text{standard}}$. This balanced sampling strategy addresses the inherent dataset bias toward straight-line driving by ensuring adequate exposure to challenging rotational motions, while preventing over-representation of any single motion pattern. \rev{The ratio favors $\mathcal{S}_{\text{high}}$ because frames on long straight segments have no high-rotation pairs. For these frames, the sampler draws from $\mathcal{S}_{\text{standard}}$, which lowers the effective share of high-rotation pairs. Appendix~\ref{appendix:hyperparam} reports the sensitivity to this ratio.}

\textbf{Temporal Consistency Benefits:} The one-minute temporal window with 4-meter spatial constraints ensures that environmental conditions, lighting characteristics, and scene content remain sufficiently consistent to maintain meaningful correspondence learning, while being large enough to include revisits of the same place within the window. For circular routes, this approach generates training pairs that simulate scenarios with moving objects, environmental changes, and significant rotational variations, effectively augmenting the dataset's diversity without requiring additional data collection efforts.

\subsection{Loss Function Design}
\label{subsec:loss_function}

The training objective combines \rev{the three pose-derived supervision signals}:
\begin{equation}
\mathcal{L}_{\text{total}} = \mathcal{L}_{\text{3-DoF}} + \lambda_1 \mathcal{L}_{\text{5-DoF}} + \lambda_2 \mathcal{L}_{\text{flow}}.
\end{equation}

The 3-DoF BEV pose loss enforces accurate planar motion estimation for ground robot applications through a weighted L1 regression:
\begin{equation}
\mathcal{L}_{\text{3-DoF}} = |t_{\text{pred},x} - t_{\text{gt},x}| + |t_{\text{pred},y} - t_{\text{gt},y}| + \alpha |\theta_{\text{pred}} - \theta_{\text{gt}}|,
\end{equation}
where $\alpha = 10$ balances translation and rotation error magnitudes to account for the different units and typical error scales in ground robot motion.

The 5-DoF PV pose loss provides additional supervisory signals for the PV branch before LSS transformation, employing a similar formulation with scale normalization. To maintain consistency with monocular constraints where absolute scale information is unavailable, we supervise only the translation direction while preserving full rotational information:
\begin{equation}
\mathcal{L}_{\text{5-DoF}} = \|\hat{\mathbf{t}}_{\text{pred}} - \hat{\mathbf{t}}_{\text{gt}}\|_1 + \beta \|\mathbf{R}_{\text{pred}} - \mathbf{R}_{\text{gt}}\|_F,
\end{equation}
where \rev{$\hat{\mathbf{t}} = \mathbf{t}/(\|\mathbf{t}\|_2+\epsilon)$} represents the unit-normalized translation vector, \rev{$\epsilon = 10^{-6}$ keeps the normalization stable for small translations,} $\|\cdot\|_F$ denotes the Frobenius norm, and $\beta = 10$. This normalization strategy ensures that scale information is exclusively handled by the depth distribution estimation and BEV projection process, enabling the PV-derived correlation cost volumes and BEV correlation cost volumes to maintain compatible feature representations for effective channel-wise concatenation and fusion. \rev{On ZJH-VO, the ground truth provides only $x$, $y$, and yaw. The roll, pitch, and vertical translation in the target are therefore set to zero, following the planar motion of this platform.}

The dense flow supervision loss enforces pixel-level correspondence learning through direct regression on the constructed ground truth flow field. \rev{It averages the end-point error (EPE) over all cells and samples:}
\begin{equation}
\rev{\mathcal{L}_{\text{flow}} = \frac{1}{N_B\,|\Omega|} \sum_{n=1}^{N_B} \sum_{q \in \Omega} \big\| \mathbf{F}_{\text{flow}}^{n}(q) - \mathbf{F}_{\text{flow-gt}}^{n}(q) \big\|_2,}
\end{equation}
\rev{where $\mathbf{F}_{\text{flow}}^{n}(q)$ and $\mathbf{F}_{\text{flow-gt}}^{n}(q)$ are the predicted and ground-truth dense rigid BEV flow at cell $q$ of sample $n$. The set $\Omega$ contains the cells of the predicted field, and $N_B$ is the batch size. Both displacement components are in grid units and weighted equally, and no normalization by motion magnitude is applied.} The direct loss formulation exploits the metric consistency of BEV representation, enabling precise pixel-level supervision without requiring additional regularization terms that could interfere with the learning of fine-grained motion patterns that support accurate correlation-based pose estimation. We set the loss weights to $\lambda_1{=}0.2$ and $\lambda_2{=}0.1$, keeping the auxiliary terms subordinate to the primary 3-DoF objective. \rev{Appendix~\ref{appendix:hyperparam} reports the sensitivity to these weights.}

\section{Experiments}

To validate our method on ground robots with different motion patterns across diverse environments, we select three public datasets (KITTI \cite{geiger2013vision}, Oxford \cite{barnes2020oxford}, and NCLT \cite{carlevaris2016university}) that range from suburban and urban driving to campus-level navigation. Additionally, we collect the ZJH-VO dataset to extend the evaluation to smaller-scale structured environments where no suitable public benchmark exists. Together, these four datasets span diverse outdoor and indoor environments with varied spatial scales and motion characteristics.

\subsection{Datasets}
\label{subsec:datasets}

\subsubsection{Public Datasets}

\textbf{KITTI Dataset:} The KITTI Odometry Dataset \cite{geiger2013vision} represents typical suburban road driving and serves as the standard benchmark for visual odometry. Following established protocols, we train on sequences 00--08 and test on 09--10. These test sequences contain significant elevation changes that challenge our planar-motion assumption.

\textbf{Oxford Dataset:} The Oxford Radar RobotCar dataset \cite{barnes2020oxford} provides complex urban driving with varying traffic density and dynamic participants. We use three training and four test sequences spanning diverse motion scales (stops, low/high speed) and spatial scales (alleys, streets). The sequence 17-12 includes strong illumination conditions, providing a challenging test for our method's generalization.

\textbf{NCLT Dataset:} The NCLT \cite{carlevaris2016university} dataset represents campus-level navigation with low-speed motion under significant environmental variations. It features significant vehicle jitter, dramatic illumination changes, and seasonal variations. Following the same split as Oxford, we train on three sequences and test on four. Notably, the sequence 08-20 features distinct seasonal conditions with foliage, testing our method's ability to generalize.

\subsubsection{ZJH-VO Multi-Scale Dataset}

BEV-based methods rely on a BEV grid with a fixed resolution, which may affect accuracy differently across scenes of different spatial scale. To test how this fixed setting holds across scales, we collect the ZJH-VO dataset on a four-wheeled mobile platform without suspension, covering four representative environments: an underground garage, an outdoor plaza, corridors, and dense offices. The dataset comprises 12 trajectories (3 per environment) with 12,666 \rev{evaluation} frames and 3,054\,m of motion data. \rev{Two additional trajectories are released for validation.} Ground truth is provided by an onboard 2D LiDAR. The platform, calibration, and ground-truth accuracy are detailed in Appendix~\ref{appendix:zjhvo}.

We use 8 trajectories (two per environment) for training and the remaining 4 for testing. Because the test trajectories differ from the training set in both timing and path coverage, \rev{they evaluate the same model on unseen trajectories from all four scene types. They also test the BEV representation under different spatial scales and motion complexity.}

\subsection{Experimental Setup}
\label{subsec:experimental_setup}

\subsubsection{Baselines}
We compare BEV-ODOM2 with a series of mainstream traditional and learning-based MVO methods, including:

\textbf{Traditional methods:} ORB-SLAM3 \cite{campos2021orb}, a classic feature-based SLAM system.

\textbf{PV learning-based methods:} These fall into two groups that match the result tables. The in-domain pose-only group requires only relative pose supervision, as ours does: DeepVO \cite{wang2017deepvo}, SSVO \cite{chen2024salient}, and VoT \cite{yugay2025visual}. The auxiliary-supervised or pretrained group requires additional supervision or external inputs: VOLDOR \cite{min2020voldor}, TartanVO \cite{wang2021tartanvo}, DF-VO \cite{zhan2021df}, DROID-SLAM \cite{teed2021droid}, and DPVO \cite{teed2024deep} rely on depth or optical flow; and MASt3R-SLAM \cite{murai2025mast3r} and VGGT-SLAM \cite{maggio2026vggt} build on pretrained 3D reconstruction priors and run zero-shot. Neither SSVO nor VoT releases official code, so we reimplement both and retrain them per dataset with pose-only supervision, matching our protocol.

\textbf{BEV learning-based methods:} BEV-DWPVO \cite{wei2025bev} and BEV-ODOM \cite{wei2024bev}. Like our current work, these methods use only pose supervision. They employ feature point extraction and matching combined with a differentiable weighted Procrustes solver, and correlation-based information extraction combined with a learning-based regression backend for pose estimation, respectively.

Because our evaluation targets real-time odometry rather than full SLAM, we disable every retrospective back-end that would revise already-emitted poses. This turns off loop closure and global bundle adjustment in all methods that include such a stage: ORB-SLAM3, DROID-SLAM, VOLDOR, MASt3R-SLAM, and VGGT-SLAM. VGGT-SLAM still infers each submap jointly over a short frame window, and we keep this design unchanged. Our method is likewise an odometry front-end, so all methods are compared at the same odometry level.

The in-domain pose-only group shares our supervision and training data and forms the main comparison. The pretrained or auxiliary-supervised group cannot be retrained on NCLT and Oxford, which provide no dense depth or optical flow ground truth. We therefore run each of these methods in its strongest feasible configuration and report them as reference points.

\subsubsection{Implementation Details and Evaluation Metrics}

We employ consistent hyperparameters across all public datasets, with BEV representation grid size of $128\times128$ and resolution of 0.8 meters, providing sufficient spatial coverage for feature mapping in BEV representation. The local correlation search range is set to $\Delta_{\text{PV}} = \Delta_{\text{BEV}} = 5$, yielding an $11\times11$ pixel correlation window in both the PV and BEV branches. The network is built using the PyTorch framework and trained on an NVIDIA RTX 4090 GPU using the Adam optimizer with an initial learning rate of $1\times10^{-4}$ for 100 epochs, with learning rate decay of 0.95 per epoch. \rev{The batch size is 16 on all datasets.}

For data augmentation, we adopt the enhanced rotation sampling strategy described in Section~\ref{subsec:enhanced_strategy}, using a 70:30 sampling ratio in favor of high-rotation pairs to ensure sufficient exposure to turning maneuvers while preventing overfitting to straight-line segments.

\rev{\textbf{Checkpoint Selection:} For BEV-ODOM2 and its ablation variants, the checkpoint is selected on held-out validation sequences used in neither training nor testing. NCLT uses the 2012-05-26 session, which lies within the period of the test sessions. Oxford uses one additional traversal from each test day. KITTI assigns all odometry sequences with ground truth to the standard split, so its validation drive comes from the raw recordings. We use 2011\_09\_26\_drive\_0022, which has the richest turning motion among the raw candidates. On these three datasets, we select the checkpoint with the lowest RTE averaged over the validation sequences. ZJH-VO uses two additional trajectories, released with the dataset, for validation. Because they are short, the selection uses their average SE(3)-aligned ATE RMSE instead of RTE.}

Evaluation metrics comprise Relative Translation Error (RTE), Relative Rotation Error (RRE), and Absolute Trajectory Error (ATE). Following standard odometry protocols, \rev{RTE and RRE are the relative translation and rotation errors averaged over all sub-trajectories of 100, 200, \ldots, 800\,m. ATE is the mean position error between the estimated and ground-truth trajectories after a single global alignment of the full sequence.} For scale consistency assessment, we employ both SE(3) and Sim(3) alignment schemes when evaluating ATE, providing insights into whether methods can maintain accurate scale without non-causal scale corrections. Notably, for MVO methods lacking absolute scale, we scale their outputs using the first 10 meters of ground truth trajectory before any evaluation, consistent with practical deployment scenarios.

\subsection{Qualitative Analysis and Case Study}

\begin{figure*}[htbp]
\centering
\includegraphics[width=1.0\textwidth]{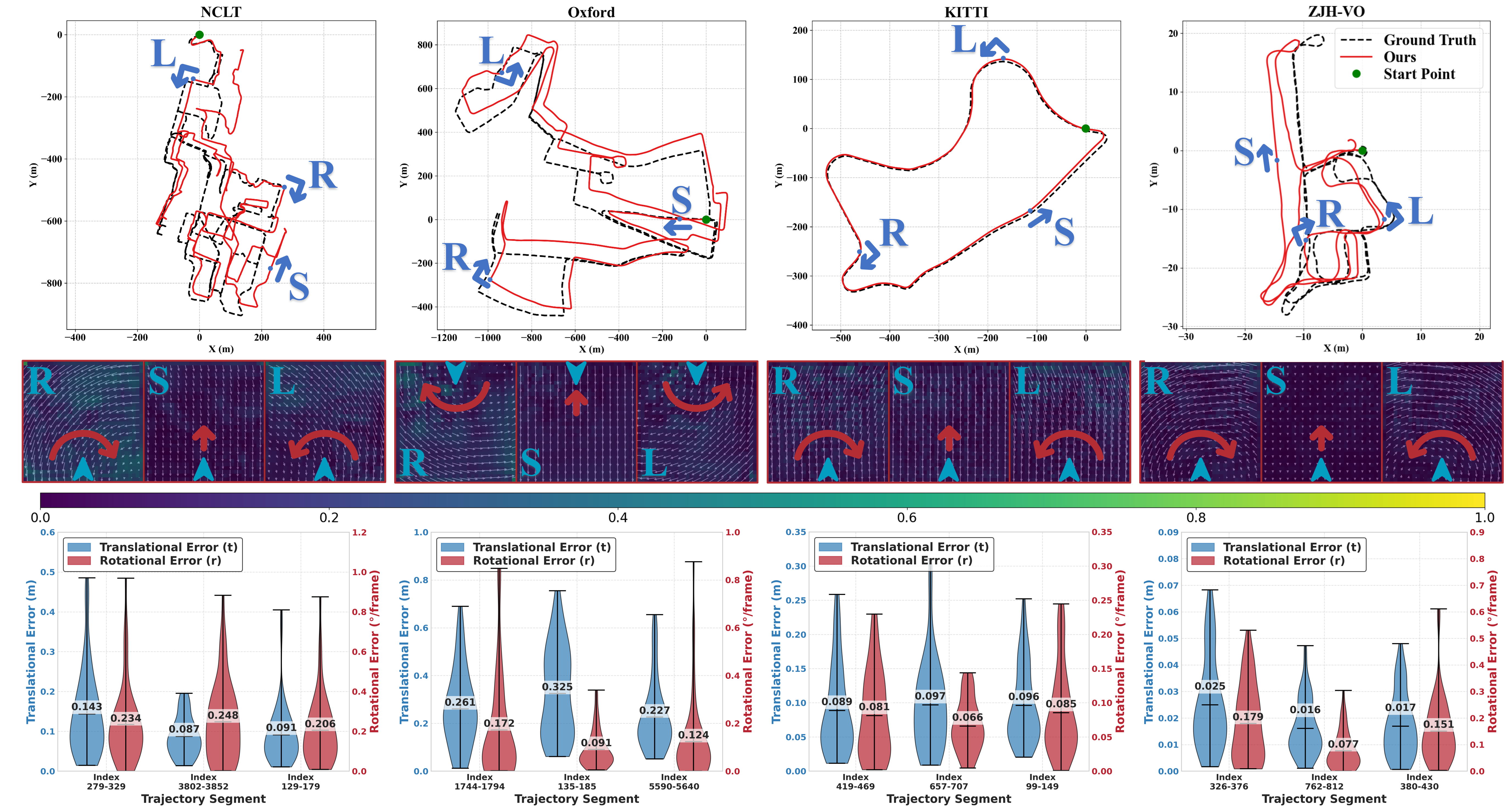}
\caption{Case Study of BEV-ODOM2 Performance on Four Datasets. The figure illustrates predicted trajectories for representative straight (S), left-turn (L), and right-turn (R) maneuvers. For each case, we visualize the predicted dense rigid BEV flow and its corresponding error map, with most pixel errors below 0.4. Below, violin plots show the distribution of translational and rotational errors for 51-frame sequences centered on these maneuvers.}
\label{fig:case_study}
\vspace{-0.4cm}
\end{figure*}  %

\subsubsection{Trajectory and Flow Visualization}
As shown in Fig.~\ref{fig:case_study}, we visualize representative sequences from BEV-ODOM2 on four datasets. The predicted trajectories (red dashed lines) closely match ground truth (black solid lines), demonstrating our method's high accuracy. We further select representative segments for right turns (R), straight driving (S), and left turns (L) from each dataset's trajectory, presenting the model's predicted dense rigid BEV flow fields and their error maps.

The BEV flow fields reveal that KITTI exhibits relatively gentle rotations, Oxford shows moderate rotations, while NCLT and ZJH-VO datasets from platforms without suspension show more dramatic rotations, consistent with expectations. These visualizations demonstrate clear motion patterns (such as arc-shaped flow fields during turns) and extremely low pixel errors (mostly below 0.4 pixels), validating the effectiveness of constructing dense supervisory signals from pose alone. The low error levels moreover confirm that such supervision produces discriminative motion-representative features, supporting more precise and more stable pose estimation. The violin plots below further illustrate inter-frame translation and rotation errors for 51-frame sequences centered on these representative segments. Thanks to enhanced rotation sampling and dense supervision, errors remain consistently small across different motion patterns for each dataset.

\begin{figure}[t]
\centering
\includegraphics[width=0.5\textwidth]{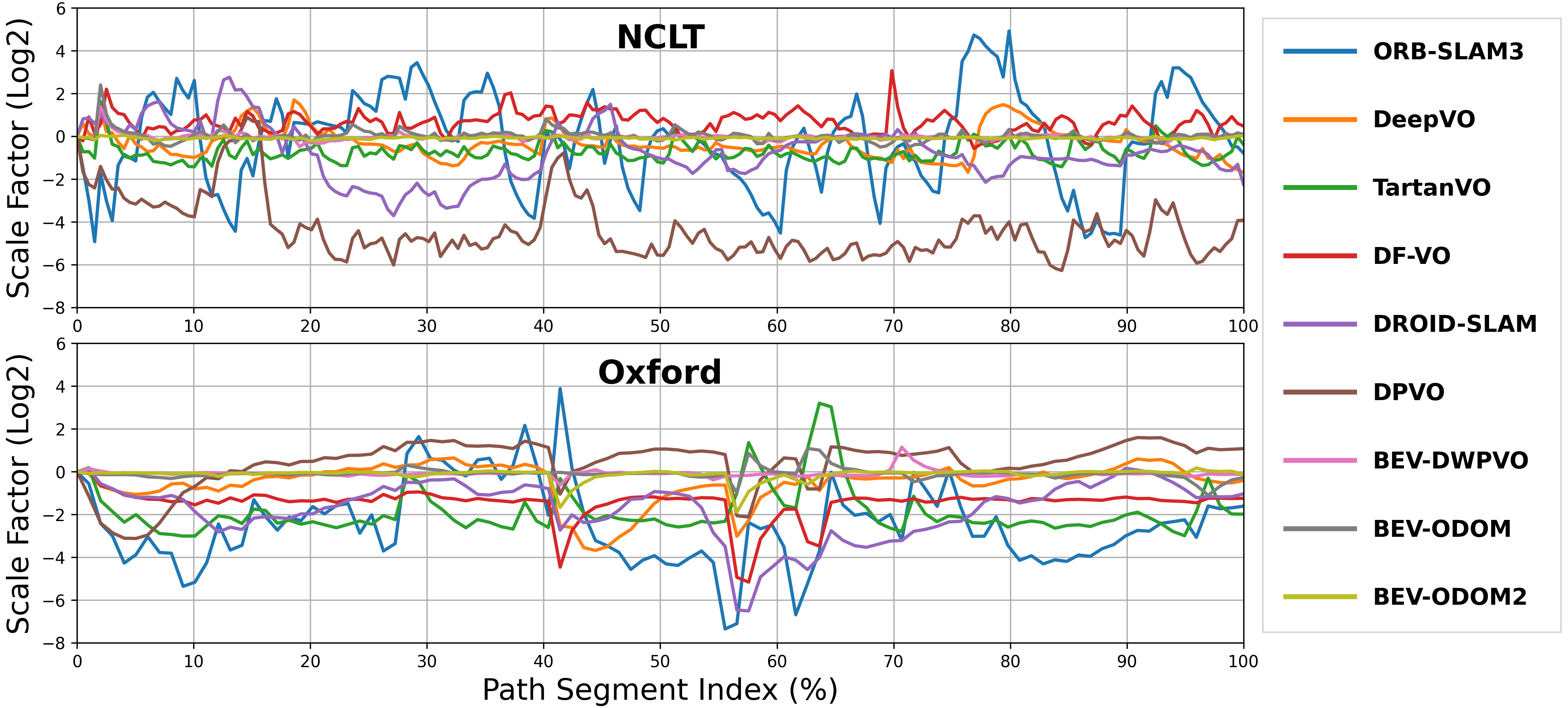}
\caption{Logarithmic scale factor variation on the NCLT (top) and Oxford (bottom) datasets, demonstrating the consistently low scale drift of BEV-based methods.}
\label{fig:scale_drift}
\vspace{-0.6cm}
\end{figure}  %

\renewcommand{\arraystretch}{1.1}  %
\begin{table*}[thbp]
    \centering
    \caption{PERFORMANCE COMPARISON ON NCLT AND OXFORD DATASETS}
    \vspace{-5pt}
    \begin{adjustbox}{max width=\textwidth}
    \begin{tabular}{
            >{\centering\arraybackslash}p{2cm}
            >{\centering\arraybackslash}p{0.9cm}
            >{\centering\arraybackslash}p{2cm}
            |>{\centering\arraybackslash}p{0.9cm}
            >{\centering\arraybackslash}p{0.9cm}
            >{\centering\arraybackslash}p{0.9cm}
            >{\centering\arraybackslash}p{0.9cm}
            >{\centering\arraybackslash}p{0.9cm}
            |>{\centering\arraybackslash}p{0.9cm}
            >{\centering\arraybackslash}p{0.9cm}
            >{\centering\arraybackslash}p{0.9cm}
            >{\centering\arraybackslash}p{0.9cm}
            >{\centering\arraybackslash}p{0.9cm}
        }
        \toprule
        \multirow{2.5}{*}{\textbf{Methods}} & \multirow{2.5}{*}{\textbf{Ex. Supv.}} & \multirow{2.5}{*}{\textbf{\diagbox[width=6em]{Metric}{Seq}}} & \multicolumn{5}{c}{\textbf{NCLT}} & \multicolumn{5}{c}{\textbf{Oxford}} \\
        \cmidrule(lr){4-8} \cmidrule(lr){9-13}
        &  &  & \textbf{02-02} & \textbf{02-19} & \textbf{03-17} & \textbf{08-20} & \textbf{\textit{Average}} & \textbf{11-12} & \textbf{15-13} & \textbf{16-14} & \textbf{17-12} & \textbf{\textit{Average}} \\
        \midrule
        \multicolumn{13}{l}{\textit{Traditional geometry-based method}} \\
        \hline
        \multirow{2}{*}{\makecell[c]{ORB-SLAM3 \cite{campos2021orb}}} &  \multirow{2}{*}{\makecell[c]{\ding{55}}} & RTE** (\%) & / & / & / & / & / & 65.22 & 952.41 & 191.75 & 77.44 & \textit{321.70} \\
                                              &  & RRE** (°/100\,m) & / & / & / & / & / & 19.83 & 10.81 & 16.17 & 19.31 & \textit{16.53} \\
        \hline
        \multicolumn{13}{l}{\textit{Pretrained or auxiliary-supervised methods (reference)}} \\
        \hline
        \multirow{2}{*}{\makecell[c]{VOLDOR \cite{min2020voldor} \\ (F. Model)}} &  \multirow{2}{*}{\makecell[c]{\ding{51}}} & RTE** (\%) & 46.76 & 41.77 & 40.39 & 45.62 & \textit{43.64} & 1123.22 & 76.02 & 142.27 & 324.06 & \textit{416.39} \\
        &  & RRE** (°/100\,m) & 17.11 & 19.26 & 26.20 & 19.26 & \textit{20.46} & 15.98 & 20.59 & 15.42 & 12.96 & \textit{16.24} \\
        \hline
        \multirow{2}{*}{\makecell[c]{TartanVO \cite{wang2021tartanvo} \\ (Pretrained Model)}} &  \multirow{2}{*}{\makecell[c]{\ding{51}}} & RTE** (\%) & 47.48 & 50.97 & 48.10 & 52.23 & \textit{49.69} & 74.43 & 83.35 & 75.51 & 72.53 & \textit{76.45} \\
                                   &  & RRE** (°/100\,m) & 39.51 & 38.48 & 39.21 & 37.68 & \textit{38.72} & 32.56 & 25.68 & 31.63 & 33.23 & \textit{30.77} \\
        \hline
        \multirow{2}{*}{\makecell[c]{DF-VO \cite{zhan2021df} \\ (Pre \& F. Model)}} &  \multirow{2}{*}{\makecell[c]{\ding{51}}} & RTE** (\%) & / & / & 41.03 & 89.44 & \textit{65.23} & 26.24 & 28.26 & 35.87 & 42.48 & \textit{33.21} \\
                                                      &  & RRE** (°/100\,m) & / & / & 25.52 & 27.81 & \textit{26.66} & 2.03 & 2.34 & 1.85 & 2.81 & \textit{2.25} \\
        \hline
        \multirow{2}{*}{\makecell[c]{DROID-SLAM \cite{teed2021droid} \\ (Pretrained Model)}} &  \multirow{2}{*}{\makecell[c]{\ding{51}}} & RTE** (\%) & 34.07 & 110.18 & 44.17 & 41.80 & \textit{57.55} & 35.18 & 136.58 & 133.31 & 25.82 & \textit{82.72} \\
                                    &  & RRE** (°/100\,m) & 14.23 & 10.50 & 10.67 & 12.04 & \textit{11.86} & 1.89 & 1.43 & 1.99 & 3.47 & \textit{2.19} \\
        \hline
        \multirow{2}{*}{\makecell[c]{DPVO \cite{teed2024deep} \\ (Pretrained Model)}}
        &  \multirow{2}{*}{\makecell[c]{\ding{51}}} & RTE** (\%) & / & 41.45 & 37.72 & 78.87 & \textit{52.68} & 96.53 & 98.41 & 102.10 & 155.51 & \textit{113.14} \\
        &  & RRE** (°/100\,m) & / & 9.74 & 14.38 & 16.70 & \textit{13.61} & 29.26 & 28.00 & 28.37 & 29.14 & \textit{28.69} \\
        \hline
        \multirow{2}{*}{\makecell[c]{MASt3R-SLAM \cite{murai2025mast3r} \\ (Pretrained Model)}} &  \multirow{2}{*}{\makecell[c]{\ding{51}}} & RTE* (\%) & / & / & / & / & / & / & / & / & / & / \\
        &  & RRE* (°/100\,m) & / & / & / & / & / & / & / & / & / & / \\
        \hline
        \multirow{2}{*}{\makecell[c]{VGGT-SLAM \cite{maggio2026vggt} \\ (Pretrained Model)}} &  \multirow{2}{*}{\makecell[c]{\ding{51}}} & RTE** (\%) & 48.56 & 50.62 & 53.64 & 68.66 & \textit{55.37} & 62.92 & 74.21 & 72.16 & 57.31 & \textit{66.65} \\
        &  & RRE** (°/100\,m) & 38.65 & 38.75 & 41.12 & 38.30 & \textit{39.21} & 14.08 & 21.26 & 24.12 & 14.57 & \textit{18.51} \\
        \hline
        \multicolumn{13}{l}{\textit{In-domain pose-only trained methods (main comparison)}} \\
        \hline
        \multirow{2}{*}{\makecell[c]{DeepVO \cite{wang2017deepvo}}} &  \multirow{2}{*}{\makecell[c]{\ding{55}}} & RTE* (\%) & 31.67 & 18.55 & 21.23 & 25.07 & \textit{24.13} & 25.14 & 27.45 & 31.75 & 37.76 & \textit{30.52} \\
                                 &  & RRE* (°/100\,m) & 14.54 & 8.11 & 9.66 & 13.02 & \textit{11.33} & 7.87 & 8.14 & 11.71 & 12.37 & \textit{10.02} \\
        \hline
        \multirow{2}{*}{\makecell[c]{SSVO \cite{chen2024salient} \\ (Our Reimpl.)}} &  \multirow{2}{*}{\makecell[c]{\ding{55}}} & RTE* (\%) & 58.00 & / & 51.31 & / & \textit{54.66} & / & / & / & / & / \\
        &  & RRE* (°/100\,m) & 41.22 & / & 43.49 & / & \textit{42.35} & / & / & / & / & / \\
        \hline
        \multirow{2}{*}{\makecell[c]{VoT \cite{yugay2025visual} \\ (Our Reimpl.)}} &  \multirow{2}{*}{\makecell[c]{\ding{55}}} & RTE* (\%) & 40.27 & 28.57 & 23.02 & 26.31 & \textit{29.54} & 70.33 & 16.61 & 70.06 & 63.79 & \textit{55.20} \\
        &  & RRE* (°/100\,m) & 19.77 & 14.34 & 13.05 & 15.70 & \textit{15.71} & 27.08 & 3.51 & 28.17 & 24.39 & \textit{20.78} \\
        \hline
        \multirow{2}{*}{\makecell[c]{BEV-DWPVO \cite{wei2025bev}}}
        &  \multirow{2}{*}{\makecell[c]{\ding{55}}} & RTE* (\%) & \underline{8.32} & 7.52 & \underline{8.20} & \underline{10.63} & \underline{\textit{8.67}} & \underline{4.87} & \underline{5.26} & \rev{\textbf{7.89}} & 9.67 & \underline{\textit{6.92}} \\
        &  & RRE* (°/100\,m) & \underline{3.14} & 3.08 & \underline{3.57} & \underline{4.35} & \underline{\textit{3.53}} & \underline{1.02} & \underline{1.08} & \underline{1.07} & \underline{2.15} & \underline{\textit{1.33}} \\
        \hline
        \multirow{2}{*}{\makecell[c]{BEV-ODOM \cite{wei2024bev}}} &  \multirow{2}{*}{\makecell[c]{\ding{55}}} & RTE* (\%) & 10.89 & \underline{6.27} & 8.77 & 13.89 & \textit{9.95} & 7.21 & 7.90 & 11.55 & \underline{8.02} & \textit{8.67} \\
                                             &  & RRE* (°/100\,m) & 4.74 & \underline{2.47} & 3.73 & 6.31 & \textit{4.31} & 1.38 & 1.89 & 1.14 & 2.55 & \textit{1.74} \\
        \hline
        \multirow{2}{*}{\makecell[c]{BEV-ODOM2 \\ (Ours)}}
        &  \multirow{2}{*}{\makecell[c]{\ding{55}}} & RTE* (\%) & \rev{\textbf{5.30}} & \rev{\textbf{4.03}} & \rev{\textbf{3.66}} & \rev{\textbf{5.75}} & \rev{\textbf{\textit{4.68}}} & \rev{\textbf{3.38}} & \rev{\textbf{3.24}} & \rev{\underline{7.98}} & \rev{\textbf{4.76}} & \rev{\textbf{\textit{4.84}}} \\
        &  & RRE* (°/100\,m) & \rev{\textbf{2.70}} & \rev{\textbf{1.62}} & \rev{\textbf{1.71}} & \rev{\textbf{2.72}} & \rev{\textbf{\textit{2.19}}} & \rev{\textbf{0.82}} & \rev{\textbf{0.87}} & \rev{\textbf{0.99}} & \rev{\textbf{1.33}} & \rev{\textbf{\textit{1.00}}} \\
        \bottomrule
    \end{tabular}
    \end{adjustbox}
    \label{table:nclt_oxford_comparison}
    \captionsetup{justification=raggedright, singlelinecheck=false}
    \caption*{\footnotesize{\textbf{Ex. Supv.}: \ding{51} indicates methods requiring additional supervision (depth, optical flow, segmentation), \ding{55} indicates pose-only supervision.\\
    \textbf{Pretrained Model}: Methods utilizing pretrained weights; \textbf{F. Model}: Foundation models for optical flow and depth estimation, i.e., Unimatch-Flow \cite{xu2023unifying} and ZoeDepth \cite{bhat2023zoedepth}. \\
    \textbf{Our Reimpl.}: No official code is released; our reimplementation, retrained per dataset with pose-only supervision.\\
    $^{*}$ Aligned using SE(3); $^{**}$ scaled by the first 10\,m of ground truth, then aligned using SE(3).}}
    \captionsetup{justification=centering, singlelinecheck=true}
\vspace{-0.8cm}
\end{table*}
\renewcommand{\arraystretch}{1.0}

\subsubsection{Scale Consistency Analysis}

Scale consistency is crucial for evaluating MVO system long-term stability, and scale drift is the dominant failure mode for monocular methods. \rev{For PV-based methods without metric scale, we scale the outputs using the ground truth over the first 10 meters, as marked in Tables~\ref{table:nclt_oxford_comparison} and~\ref{table:kitti_zjh}.} As shown in Fig.~\ref{fig:scale_drift}, we plot the logarithmic scale factor variation curves for different methods on the challenging NCLT and Oxford datasets, calculated as:
\begin{equation}
s_i = \log_2\left(\frac{d_i}{d_{i}^{\text{GT}}}\right),
\end{equation}
where \( d_i \) and \( d_{i}^{\text{GT}} \) represent the estimated and ground truth displacements for segment \( i \). The logarithmic space representation naturally accommodates the multiplicative characteristics of scale drift in MVO systems. The resulting curves illustrate how scale consistency evolves across different trajectory segments, providing insights into the long-term drift behavior of each method.

Compared to most methods, all BEV-based approaches, including BEV-DWPVO, BEV-ODOM, and BEV-ODOM2, demonstrate scale factor curves that remain closer to zero with significantly reduced fluctuations. This performance validates the effectiveness of the implicit scale anchoring strategy achieved through the BEV grid representation, which has been consistently adopted throughout the BEV-ODOM line. Notably, these methods achieve excellent scale consistency using only pose supervision without requiring extra supervisory signals.

Despite incorporating a more complex PV-BEV fusion architecture, BEV-ODOM2 successfully preserves this superior scale consistency characteristic. This preservation is attributed to the unified processing pipeline employed by both motion feature extraction pathways. Specifically, the BEV motion features, obtained through projection followed by correlation, and the PV motion features, derived from correlation followed by projection, both utilize the same depth estimation network and BEV representation projection mechanism. This shared pipeline ensures that all motion features are anchored to a consistent metric scale defined by the BEV grid structure, thereby maintaining scale consistency throughout the entire framework.

\subsection{Quantitative Analysis}

\renewcommand{\arraystretch}{1.1}  %
\begin{table*}[thbp]
    \centering
    \caption{PERFORMANCE COMPARISON ON KITTI AND ZJH-VO DATASETS}
    \vspace{-5pt}
    \begin{adjustbox}{max width=\textwidth}
    \begin{tabular}{
            >{\centering\arraybackslash}c
            >{\centering\arraybackslash}c
            |>{\centering\arraybackslash}c
            >{\centering\arraybackslash}c
            >{\centering\arraybackslash}c
            |>{\centering\arraybackslash}c
            >{\centering\arraybackslash}c
            >{\centering\arraybackslash}c
            >{\centering\arraybackslash}c
            >{\centering\arraybackslash}c
        }
        \toprule
        \multirow{2.5}{*}{\textbf{Methods}} & \multirow{2.5}{*}{\textbf{\diagbox[width=6em]{Metric}{Seq}}} & \multicolumn{3}{c}{\textbf{KITTI}} & \multicolumn{5}{c}{\textbf{ZJH-VO}} \\
        \cmidrule(lr){3-5} \cmidrule(lr){6-10}
        & & \textbf{09} & \textbf{10} & \textbf{\textit{Average}} & \textbf{F-09} & \textbf{F-04} & \textbf{F-01} & \textbf{F-00} & \textbf{\textit{Average}} \\
        \midrule
        \multicolumn{10}{l}{\textit{Traditional geometry-based method}} \\
        \hline
        \multirow{4}{*}{\makecell[c]{ORB-SLAM3 \cite{campos2021orb}}}
            & RTE** (\%) & 14.89 & 8.47 & \textit{11.68} & \textcolor{gray}{9.63} & \textcolor{gray}{14.76} & \textcolor{gray}{4.57} & \textcolor{gray}{6.77} & \textcolor{gray}{\textit{8.94}} \\
            & RRE** (°/100\,m) & 1.38 & 0.93 & \textit{1.15} & \textcolor{gray}{13.67} & \textcolor{gray}{28.26} & \textcolor{gray}{6.95} & \textcolor{gray}{7.97} & \textcolor{gray}{\textit{14.21}} \\
            & ATE** (m) & 53.69 & 16.15 & \textit{34.92} & \textcolor{gray}{6.34} & \textcolor{gray}{5.73} & \textcolor{gray}{2.03} & \textcolor{gray}{3.00} & \textcolor{gray}{\textit{4.28}} \\
            & ATE$^{\dag}$ (m) & 40.63 & 7.33 & \textit{23.98} & \textcolor{gray}{6.12} & \textcolor{gray}{5.54} & \textcolor{gray}{1.91} & \textcolor{gray}{1.66} & \textcolor{gray}{\textit{3.81}} \\
        \hline
        \multicolumn{10}{l}{\textit{Pretrained or auxiliary-supervised methods (reference)}} \\
        \hline
        \multirow{4}{*}{\makecell[c]{VOLDOR \cite{min2020voldor} \\ (F. Model)}}
            & RTE** (\%) & 58.45 & 68.64 & \textit{63.55} & 14.76 & 8.13 & 38.23 & 35.88 & \textit{24.25} \\
            & RRE** (°/100\,m) & 0.60 & 0.77 & \textit{0.69} & 36.33 & 84.80 & 34.04 & 28.91 & \textit{46.02} \\
            & ATE** (m) & \rev{168.83} & \rev{158.81} & \rev{\textit{163.82}} & \rev{11.61} & \rev{11.60} & \rev{24.79} & \rev{24.29} & \rev{\textit{18.07}} \\
            & ATE$^{\dag}$ (m) & \rev{152.98} & \rev{125.78} & \rev{\textit{139.38}} & \rev{11.63} & \rev{11.23} & \rev{20.54} & \rev{24.45} & \rev{\textit{16.96}} \\
        \hline
        \multirow{4}{*}{\makecell[c]{TartanVO \cite{wang2021tartanvo} \\ {\scriptsize KITTI: reported} \\ {\scriptsize ZJH: retrained}}}
            & RTE* (\%) & 6.00 & 6.89 & \textit{6.45} & 5.98 & 5.27 & 23.27 & 6.71 & \textit{10.31} \\
            & RRE* (°/100\,m) & 3.11 & 2.73 & \textit{2.92} & 14.76 & 18.46 & 29.79 & 6.58 & \textit{17.40} \\
            & ATE* (m) & 53.84 & 28.50 & \textit{41.17} & 3.52 & 3.16 & 14.00 & 2.48 & \textit{5.79} \\
            & ATE$^{\dag}$ (m) & 27.52 & 23.65 & \textit{25.59} & 3.48 & 3.14 & 11.70 & 2.19 & \textit{5.13} \\
        \hline
        \multirow{4}{*}{\makecell[c]{DF-VO \cite{zhan2021df} \\ {\scriptsize KITTI: reported} \\ {\scriptsize ZJH: Pre \& F. Model}}}
            & RTE$^{*/**}$ (\%) & 2.40 & \textbf{1.82} & \textbf{\textit{2.11}} & 31.31 & 105.42 & 55.16 & 48.11 & \textit{60.00} \\
            & RRE$^{*/**}$ (°/100\,m) & \textbf{0.24} & 0.38 & \textit{0.31} & 88.26 & 141.32 & 117.93 & 75.13 & \textit{105.66} \\
            & ATE$^{*/**}$ (m) & 7.73 & \textbf{3.00} & \underline{\textit{5.37}} & 116.79 & 64.45 & 45.99 & 19.71 & \textit{61.73} \\
            & ATE$^{\dag}$ (m) & 7.65 & \textbf{2.73} & \underline{\textit{5.19}} & 10.56 & 11.96 & 24.02 & 18.94 & \textit{16.37} \\
        \hline
        \multirow{4}{*}{\makecell[c]{DROID-SLAM \cite{teed2021droid} \\ (Pretrained Model)}}
            & RTE** (\%) & 21.01 & 18.73 & \textit{19.87} & 16.85 & 6.98 & 14.15 & 25.00 & \textit{15.75} \\
            & RRE** (°/100\,m) & 0.33 & \textbf{0.23} & \underline{\textit{0.28}} & 25.50 & 26.36 & 6.15 & 30.44 & \textit{22.11} \\
            & ATE** (m) & 73.71 & 45.52 & \textit{59.61} & 12.52 & 8.91 & 8.82 & 17.00 & \textit{11.81} \\
            & ATE$^{\dag}$ (m) & 74.31 & 17.27 & \textit{45.79} & 11.65 & 8.33 & \rev{3.57} & 10.78 & \textit{8.58} \\
        \hline
        \multirow{4}{*}{\makecell[c]{DPVO \cite{teed2024deep} \\ (Pretrained Model)}}
            & RTE** (\%) & 17.70 & 4.47 & \textit{11.09} & 14.67 & 6.55 & 11.08 & 26.65 & \textit{14.74} \\
            & RRE** (°/100\,m) & \textbf{0.24} & \textbf{0.23} & \textbf{\textit{0.23}} & 53.48 & \textbf{3.33} & 4.37 & 17.42 & \textit{19.65} \\
            & ATE** (m) & 64.80 & 10.98 & \textit{37.89} & 11.11 & 7.46 & 6.94 & 18.08 & \textit{10.90} \\
            & ATE$^{\dag}$ (m) & 64.77 & 10.99 & \textit{37.88} & 11.10 & 6.08 & \textbf{3.20} & 12.29 & \textit{8.16} \\
        \hline
        \multirow{4}{*}{\makecell[c]{MASt3R-SLAM \cite{murai2025mast3r} \\ (Pretrained Model)}}
            & RTE* (\%) & / & / & / & 14.38 & 8.40 & 46.39 & 34.97 & \textit{26.03} \\
            & RRE* (°/100\,m) & / & / & / & 8.48 & 9.18 & 9.78 & 14.44 & \textit{10.47} \\
            & ATE* (m) & / & / & / & \rev{11.32} & \rev{11.06} & \rev{23.51} & \rev{23.19} & \rev{\textit{17.27}} \\
            & ATE$^{\dag}$ (m) & / & / & / & \rev{11.38} & \rev{10.03} & \rev{16.00} & \rev{22.60} & \rev{\textit{15.00}} \\
        \hline
        \multirow{4}{*}{\makecell[c]{VGGT-SLAM \cite{maggio2026vggt} \\ (Pretrained Model)}}
            & RTE** (\%) & 22.90 & 14.50 & \textit{18.70} & 8.84 & 5.39 & \underline{8.29} & 15.37 & \textit{9.47} \\
            & RRE** (°/100\,m) & 2.86 & 3.06 & \textit{2.96} & 32.77 & 16.81 & \textbf{3.77} & 5.01 & \textit{14.59} \\
            & ATE** (m) & \rev{72.35} & \rev{20.39} & \rev{\textit{46.37}} & \rev{6.67} & \rev{4.06} & \rev{\underline{5.46}} & \rev{10.69} & \rev{\textit{6.72}} \\
            & ATE$^{\dag}$ (m) & \rev{72.47} & \rev{19.88} & \rev{\textit{46.18}} & \rev{6.10} & \rev{2.68} & \rev{\underline{3.24}} & \rev{3.94} & \rev{\textit{3.99}} \\
        \hline
        \multicolumn{10}{l}{\textit{In-domain pose-only trained methods (main comparison)}} \\
        \hline
        \multirow{4}{*}{\makecell[c]{DeepVO \cite{wang2017deepvo} \\ (Retrained)}}
            & RTE* (\%) & 33.55 & 30.46 & \textit{32.01} & 51.05 & 37.52 & 56.08 & 63.16 & \textit{51.95} \\
            & RRE* (°/100\,m) & 14.31 & 14.42 & \textit{14.37} & 106.52 & 116.07 & 102.27 & 111.58 & \textit{109.11} \\
            & ATE* (m) & 59.38 & 180.10 & \textit{119.74} & 25.65 & 18.71 & 14.72 & 21.17 & \textit{20.06} \\
            & ATE$^{\dag}$ (m) & 51.58 & 168.44 & \textit{110.01} & 10.79 & 10.04 & 14.36 & 15.81 & \textit{12.75} \\
        \hline
        \multirow{4}{*}{\makecell[c]{SSVO \cite{chen2024salient} \\ (Our Reimpl.)}}
            & RTE* (\%) & 71.47 & 78.57 & \textit{75.02} & / & / & / & / & / \\
            & RRE* (°/100\,m) & 0.67 & 0.39 & \textit{0.53} & / & / & / & / & / \\
            & ATE* (m) & \rev{203.34} & \rev{174.77} & \rev{\textit{189.05}} & / & / & / & / & / \\
            & ATE$^{\dag}$ (m) & \rev{30.98} & \rev{\underline{5.60}} & \rev{\textit{18.29}} & / & / & / & / & / \\
        \hline
        \multirow{4}{*}{\makecell[c]{VoT \cite{yugay2025visual} \\ (Our Reimpl.)}}
            & RTE* (\%) & 91.02 & 85.11 & \textit{88.06} & 2.59 & 4.27 & 24.74 & 8.64 & \textit{10.06} \\
            & RRE* (°/100\,m) & 22.28 & 34.16 & \textit{28.22} & 8.31 & 13.02 & 35.55 & 13.21 & \textit{17.52} \\
            & ATE* (m) & \rev{106.88} & \rev{42.66} & \rev{\textit{74.77}} & \rev{1.38} & \rev{2.73} & \rev{14.87} & \rev{3.93} & \rev{\textit{5.73}} \\
            & ATE$^{\dag}$ (m) & \rev{79.89} & \rev{36.76} & \rev{\textit{58.32}} & \rev{1.29} & \rev{2.68} & \rev{13.48} & \rev{2.78} & \rev{\textit{5.06}} \\
        \hline
        \multirow{4}{*}{\makecell[c]{BEV-DWPVO \cite{wei2025bev}}}
            & RTE* (\%) & 2.13 & \underline{3.24} & \textit{2.69} & 2.75 & 3.74 & 16.62 & 5.50 & \underline{\textit{7.15}} \\
            & RRE* (°/100\,m) & 0.65 & 1.15 & \textit{0.90} & 7.18 & 9.28 & 12.72 & \rev{\textbf{1.98}} & \underline{\textit{7.79}} \\
            & ATE* (m) & 8.69 & 9.04 & \textit{8.87} & \rev{1.75} & \rev{2.42} & \rev{10.68} & \rev{3.39} & \rev{\textit{4.56}} \\
            & ATE$^{\dag}$ (m) & 8.11 & 7.80 & \textit{7.96} & \rev{1.67} & \rev{2.40} & \rev{6.34} & \rev{2.54} & \rev{\underline{\textit{3.24}}} \\
        \hline
        \multirow{4}{*}{\makecell[c]{BEV-ODOM \cite{wei2024bev}}}
            & RTE* (\%) & \underline{1.72} & 3.61 & \textit{2.67} & \rev{\underline{1.91}} & \underline{3.47} & 20.29 & \underline{3.46} & \textit{7.28} \\
            & RRE* (°/100\,m) & 0.39 & 0.53 & \textit{0.46} & \underline{6.39} & \underline{7.83} & 13.28 & \rev{\underline{4.41}} & \textit{7.98} \\
            & ATE* (m) & \underline{6.35} & 8.42 & \textit{7.39} & \rev{\underline{1.15}} & \underline{2.00} & 11.00 & \underline{1.63} & \underline{\textit{3.95}} \\
            & ATE$^{\dag}$ (m) & \underline{4.62} & 7.30 & \textit{5.96} & \rev{\underline{1.12}} & \underline{1.91} & 10.94 & \rev{\textbf{0.77}} & \textit{3.68} \\
        \hline
        \multirow{4}{*}{\makecell[c]{BEV-ODOM2 \\ (Ours)}}
            & RTE* (\%) & \textbf{1.37} & 3.46 & \underline{\textit{2.42}} & \rev{\textbf{1.67}} & \rev{\textbf{3.29}} & \rev{\textbf{6.57}} & \rev{\textbf{3.08}} & \rev{\textbf{\textit{3.65}}} \\
            & RRE* (°/100\,m) & 0.35 & 0.52 & \textit{0.44} & \rev{\textbf{4.59}} & \rev{8.48} & \rev{\underline{3.88}} & \rev{5.38} & \rev{\textbf{\textit{5.58}}} \\
            & ATE* (m) & \textbf{2.89} & \underline{6.08} & \textbf{\textit{4.49}} & \rev{\textbf{1.01}} & \rev{\textbf{1.83}} & \rev{\textbf{3.69}} & \rev{\textbf{1.58}} & \rev{\textbf{\textit{2.03}}} \\
            & ATE$^{\dag}$ (m) & \textbf{2.72} & \rev{5.97} & \textbf{\textit{4.35}} & \rev{\textbf{1.00}} & \rev{\textbf{1.82}} & \rev{3.43} & \rev{\underline{1.30}} & \rev{\textbf{\textit{1.89}}} \\
    \bottomrule
    \end{tabular}
    \end{adjustbox}
    \label{table:kitti_zjh}
    \captionsetup{justification=raggedright, singlelinecheck=false}
    \caption*{\footnotesize{Method configuration is shown as KITTI\,/\,ZJH-VO where it differs. \textbf{reported}: results from the original publication; \textbf{Retrained}: retrained in-domain; \textbf{Pretrained Model}: evaluated from pretrained weights; \textbf{F. Model}: foundation models for optical flow and depth. \textbf{Our Reimpl.}: no official code is released; our reimplementation, retrained per dataset with pose-only supervision. Gray entries indicate ORB-SLAM3 completes only partial ZJH-VO sequences. \\
    $^{*}$ Aligned using SE(3); $^{**}$ scaled by the first 10\,m of ground truth, then aligned using SE(3); $^{\dag}$ aligned using Sim(3).}}
    \captionsetup{justification=centering, singlelinecheck=true}
\vspace{-0.4cm}
\end{table*}
\renewcommand{\arraystretch}{1.0}

\subsubsection{Performance on NCLT and Oxford Datasets}

As shown in Table~\ref{table:nclt_oxford_comparison}, BEV-ODOM2 demonstrates superior performance on the challenging NCLT and Oxford datasets, achieving RTE of \rev{4.68\% and 4.84\%} with RRE of \rev{2.19°/100\,m and 1.00°/100\,m}, respectively.

The traditional ORB-SLAM3 method fails completely on NCLT due to feature tracking loss in high-jitter environments and suffers from severe scale drift on Oxford with numerous dynamic objects, highlighting the inherent challenges of these datasets. Among perspective-view methods, VOLDOR, TartanVO, DF-VO, DROID-SLAM, and DPVO all depend on optical flow or depth. Neither dataset provides the ground truth to retrain or fine-tune these components, so their translation errors remain several times to more than ten times higher than ours. MASt3R-SLAM does not obtain complete trajectories on these kilometer-scale sequences. VGGT-SLAM processes the long sequences as fixed-size submaps, and its errors also remain far above ours. Among methods using pose-only supervision like ours, DeepVO and VoT demonstrate inferior performance compared to the BEV-based approaches, and SSVO does not obtain complete trajectories on most sequences.

Within BEV methods, our approach provides denser supervision compared to BEV-ODOM and offers more flexible constraints compared to BEV-DWPVO's solver while incorporating PV branch information through correlation-based fusion. This leads to notable improvements over BEV-DWPVO, with RTE reductions of \rev{46.0\%} on NCLT and \rev{30.1\%} on Oxford, and RRE reductions of \rev{38.0\%} and \rev{24.8\%} on the two datasets, respectively.

\subsubsection{Performance on KITTI Dataset}
\label{subsubsec:kitti_results}

As presented in Table~\ref{table:kitti_zjh}, BEV-ODOM2 achieves the best RTE and ATE on the loop sequence 09, and an average ATE of 4.49\,m under SE(3) alignment without any ground truth scale, surpassing most baselines that use Sim(3) alignment with scale correction. The gap to the in-domain pose-only PV methods comes from scale. DeepVO, SSVO, and VoT are trained with the same pose-only supervision as ours, yet their RTE stays above 30\% because monocular scale drifts without a metric reference. The metric BEV grid removes this ambiguity by construction, so the BEV methods keep RTE near 2--3\% under the identical protocol. On sequence 10, which contains significant elevation changes that violate the planar motion assumption, our RTE is higher than DF-VO (trained with stereo depth and flow supervision), yet the overall accuracy remains competitive.

\begin{figure}[t]
\centering
\includegraphics[width=0.49\textwidth]{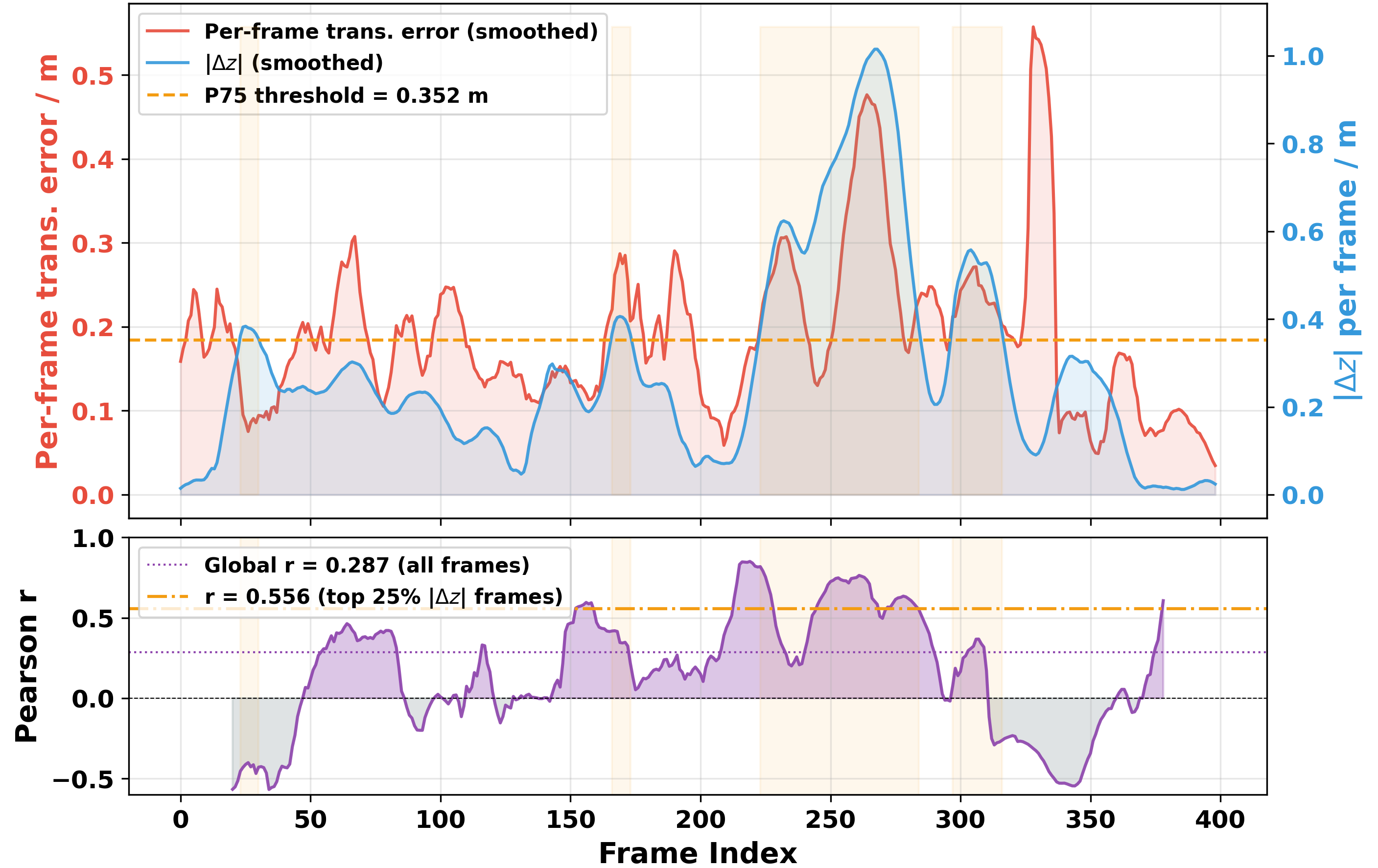}
\caption{Sensitivity analysis on KITTI Seq10. \textbf{Top}: \rev{per-frame translation error in meters} and inter-frame elevation change $|\Delta z|$, with the P75 threshold marking high-elevation-change regions. \textbf{Bottom}: sliding-window Pearson correlation between $|\Delta z|$ and \rev{the per-frame translation error}.}
\label{fig:elevation_analysis}
\vspace{-0.3cm}
\end{figure}

We quantify this elevation effect directly. Fig.~\ref{fig:elevation_analysis} plots the \rev{per-frame translation error} on sequence 10 alongside the inter-frame elevation change $|\Delta z|$ from the 6-DoF ground truth. \rev{This error is the norm of the difference between the estimated and ground-truth inter-frame horizontal displacements, measured in meters. Unlike the RTE, it is not averaged over fixed-length sub-trajectories. The Pearson correlation is $r{=}0.287$ over all frames and rises to $r{=}0.556$ in the top 25\% high-$|\Delta z|$ frames. This suggests that elevation change is an important contributing factor to the error on this sequence, particularly in segments with large elevation changes.} The full model still outperforms BEV-ODOM on both KITTI sequences, but the error induced by elevation changes remains an inherent limitation of the 3-DoF formulation.

\begin{figure}[t]
\centering
\includegraphics[width=0.49\textwidth]{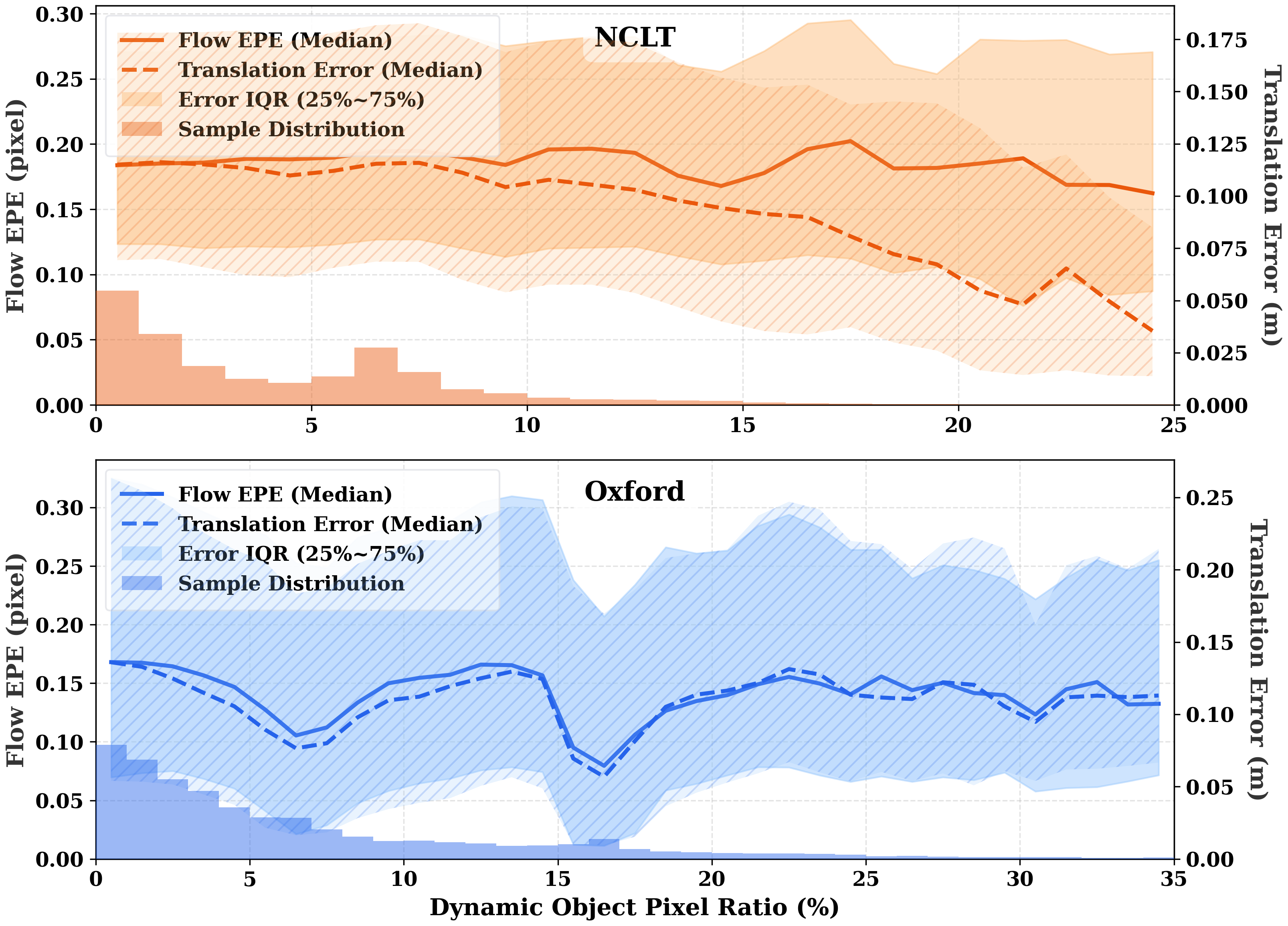}
\caption{Impact of dynamic objects on estimation accuracy. Each panel shows the median BEV flow end-point error (solid) and \rev{median per-frame translation error in meters (dashed)} versus the per-frame dynamic object pixel ratio, with interquartile ranges shaded and sample counts shown as histograms.}
\label{fig:dynamic_error}
\vspace{-0.6cm}
\end{figure}

\subsubsection{Performance on ZJH-VO Dataset}

As shown in Table~\ref{table:kitti_zjh}, the advantages of BEV-ODOM2 are further validated on the ZJH-VO multi-scale dataset, which spans both outdoor and indoor environments with diverse spatial scales. It achieves the best performance on the average metrics \rev{over the four scene types}. Compared to BEV-ODOM using only sparse pose supervision, the dense rigid BEV flow supervision provides richer supervisory signals, enabling the network to learn clearer mappings of objects at different distances in BEV representation during training. This enhanced feature representation not only improves pose estimation accuracy but also maintains exceptional scale consistency. Despite the challenging multi-scale environment that tests BEV grid representation limits, BEV-ODOM2 achieves superior ATE under SE(3) alignment compared to most methods using ground truth scale correction.

The baselines are most competitive on these short-range sequences. VGGT-SLAM and VoT reach their lowest RTE among the four datasets here, yet neither leads on any average metric. BEV-ODOM2 reaches an RTE of \rev{3.65\%} and a Sim(3) ATE of \rev{1.89\,m}.

\renewcommand{\arraystretch}{0.925}
\begin{table}[t]
\centering
\caption{COMPUTATIONAL COST COMPARISON UNDER TWO INPUT RESOLUTIONS}
\vspace{-5pt}
\begin{adjustbox}{max width=\columnwidth}
\begin{tabular}{cc|ccc}
\toprule
Method & Resolution & FPS & \makecell{GPU Mem \\ (GB)} & \makecell{GPU Util \\ (\%)} \\
\midrule
\multicolumn{5}{l}{\textit{Traditional geometry-based method}} \\
\midrule
\multirow{2}{*}{ORB-SLAM3 \cite{campos2021orb}} & 224$\times$384 & 63.11 & - & - \\
 & 320$\times$640 & 59.68 & - & - \\
\midrule
\multicolumn{5}{l}{\textit{Pretrained or auxiliary-supervised methods (reference)}} \\
\midrule
\multirow{2}{*}{VOLDOR \cite{min2020voldor}} & 224$\times$384 & 17.27 & 1.27 & 63.0 \\
 & 320$\times$640 & 12.30 & 1.83 & 71.7 \\
\midrule
\multirow{2}{*}{TartanVO \cite{wang2021tartanvo}} & 224$\times$384 & 63.45 & 2.026 & 56.0 \\
 & 320$\times$640 & 59.27 & 2.155 & 60.9 \\
\midrule
\multirow{2}{*}{DF-VO \cite{zhan2021df}} & 224$\times$384 & 9.37 & 3.037 & - \\
 & 320$\times$640 & 6.26 & 3.291 & - \\
\midrule
\multirow{2}{*}{DROID-SLAM \cite{teed2021droid}} & 224$\times$384 & 10.58 & 21.138 & 54.7 \\
 & 320$\times$640 & 7.30 & 21.231 & 54.0 \\
\midrule
\multirow{2}{*}{DPVO \cite{teed2024deep}} & 224$\times$384 & 18.88 & 4.226 & 65.9 \\
 & 320$\times$640 & 18.03 & 4.251 & 68.7 \\
\midrule
\multirow{2}{*}{MASt3R-SLAM \cite{murai2025mast3r}} & 224$\times$384 & 12.62 & 4.703 & 78.7 \\
 & 320$\times$640 & 12.62 & 4.703 & 78.7 \\
\midrule
\multirow{2}{*}{VGGT-SLAM \cite{maggio2026vggt}} & 224$\times$384 & 8.36 & 14.259 & 41.6 \\
 & 320$\times$640 & 8.36 & 14.259 & 41.6 \\
\midrule
\multicolumn{5}{l}{\textit{In-domain pose-only trained methods (main comparison)}} \\
\midrule
\multirow{2}{*}{DeepVO \cite{wang2017deepvo}} & 224$\times$384 & 416.70 & 3.176 & 41.3 \\
 & 320$\times$640 & 348.24 & 4.366 & 38.5 \\
\midrule
\multirow{2}{*}{SSVO \cite{chen2024salient}} & 224$\times$384 & 35.78 & 1.785 & 71.2 \\
 & 320$\times$640 & 35.51 & 1.996 & 74.4 \\
\midrule
\multirow{2}{*}{VoT \cite{yugay2025visual}} & 224$\times$384 & 66.90 & 2.718 & 59.7 \\
 & 320$\times$640 & 25.04 & 3.216 & 87.2 \\
\midrule
\multirow{2}{*}{BEV-DWPVO \cite{wei2025bev}} & 224$\times$384 & 21.61 & 2.566 & 40.0 \\
 & 320$\times$640 & 18.21 & 3.138 & 49.2 \\
\midrule
\multirow{2}{*}{BEV-ODOM \cite{wei2024bev}} & 224$\times$384 & 92.84 & 2.289 & 59.7 \\
 & 320$\times$640 & 72.64 & 2.503 & 68.7 \\
\midrule
\multirow{2}{*}{\makecell{BEV-ODOM2 \\ (Ours)}} & 224$\times$384 & 69.52 & 2.304 & 54.4 \\
 & 320$\times$640 & 56.62 & 2.525 & 64.3 \\
\midrule
\multicolumn{5}{c}{\textit{Edge Deployment on NVIDIA Jetson AGX Orin}} \\
\midrule
\multirow{2}{*}{\makecell{BEV-ODOM2 \\ (FP32)}} & 224$\times$384 & 16.91 & 1.625 & 63.7 \\
 & 320$\times$640 & 14.14 & 1.650 & 76.7 \\
\midrule
\multirow{2}{*}{\makecell{BEV-ODOM2 \\ (TRT Backbone)}} & 224$\times$384 & 21.84 & 1.576 & 66.2 \\
 & 320$\times$640 & 16.73 & 1.600 & 78.0 \\
\bottomrule
\end{tabular}
\end{adjustbox}
\label{table:computational_cost}
\captionsetup{justification=raggedright, singlelinecheck=false}
\caption*{\footnotesize{All learning-based methods are evaluated on an RTX 4090 with batch size 1; ORB-SLAM3 runs on CPU (Intel Core i7-12700KF). MASt3R-SLAM and VGGT-SLAM resize inputs to a fixed internal resolution. Edge deployment rows are measured on an NVIDIA Jetson AGX Orin, where TRT Backbone converts only the ResNet-50 to TensorRT FP16.}}
\captionsetup{justification=centering, singlelinecheck=true}
\vspace{-0.8cm}
\end{table}
\renewcommand{\arraystretch}{1.0}

\begin{table*}[t]
\centering
\caption{CLOSED-LOOP ROUTE-FOLLOWING PERFORMANCE ON THE PHYSICAL ROBOT}
\vspace{-4pt}
\begin{adjustbox}{max width=\textwidth}
\begin{tabular}{ccc ccc ccc}
\toprule
\multirow{2}{*}{Route} & \multirow{2}{*}{Length (m)} & \multirow{2}{*}{Method} & \multicolumn{3}{c}{Closed-loop task (actual vs.\ reference)} & \multicolumn{3}{c}{Odometry accuracy (VO vs.\ LiDAR GT)} \\
\cmidrule(lr){4-6} \cmidrule(lr){7-9}
 & & & Coll.-free dist.\ (m) & Lat.\ RMSE (cm) & Term.\ err.\ (cm) & ATE (cm) & RTE (\%) & RRE (°/10\,m) \\
\midrule
\multirow{2}{*}{T} & \multirow{2}{*}{18.1} & BEV-ODOM & 13.2\,$\pm$\,3.3 & 18.2\,$\pm$\,4.5 & 23.1\,$\pm$\,16.6 & 32.0\,$\pm$\,6.4 & 7.3\,$\pm$\,2.2 & \textbf{1.54\,$\pm$\,0.56} \\
 & & \textbf{BEV-ODOM2} & \textbf{18.1}~(all~\ding{51}) & \textbf{10.3\,$\pm$\,3.6} & \textbf{21.7\,$\pm$\,14.4} & \textbf{14.4\,$\pm$\,4.8} & \textbf{2.6\,$\pm$\,1.0} & 1.72\,$\pm$\,0.68 \\
\midrule
\multirow{2}{*}{A} & \multirow{2}{*}{21.3} & BEV-ODOM & 19.7\,$\pm$\,2.3 & 32.0\,$\pm$\,5.3 & 73.9\,$\pm$\,16.9 & 33.9\,$\pm$\,5.0 & 5.2\,$\pm$\,1.0 & 3.69\,$\pm$\,1.68 \\
 & & \textbf{BEV-ODOM2} & \textbf{21.3}~(all~\ding{51}) & \textbf{12.8\,$\pm$\,3.7} & \textbf{23.6\,$\pm$\,9.9} & \textbf{18.4\,$\pm$\,3.6} & \textbf{2.8\,$\pm$\,0.5} & \textbf{1.46\,$\pm$\,0.43} \\
\midrule
\multirow{2}{*}{S} & \multirow{2}{*}{12.9} & BEV-ODOM & 5.3\,$\pm$\,1.6 & -- & -- & -- & 18.8\,$\pm$\,1.6 & 5.99\,$\pm$\,1.48 \\
 & & \textbf{BEV-ODOM2} & \textbf{12.9}~(all~\ding{51}) & \textbf{7.4\,$\pm$\,1.4} & \textbf{8.4\,$\pm$\,6.1} & \textbf{13.8\,$\pm$\,1.5} & \textbf{2.8\,$\pm$\,0.9} & \textbf{1.15\,$\pm$\,0.54} \\
\midrule
\multirow{2}{*}{E} & \multirow{2}{*}{23.1} & BEV-ODOM & 19.3\,$\pm$\,0.4 & -- & -- & -- & 6.3\,$\pm$\,1.4 & 6.65\,$\pm$\,2.13 \\
 & & \textbf{BEV-ODOM2} & \textbf{22.4\,$\pm$\,1.8} & \textbf{15.1\,$\pm$\,3.4} & \textbf{29.0\,$\pm$\,29.2} & \textbf{21.2\,$\pm$\,4.1} & \textbf{4.3\,$\pm$\,2.0} & \textbf{4.67\,$\pm$\,0.49} \\
\bottomrule
\end{tabular}
\end{adjustbox}
\label{table:closedloop_success}
\captionsetup{justification=raggedright, singlelinecheck=false}
\caption*{\footnotesize{Each route is run 10 times per method. \emph{Coll.-free dist.}\ is the collision-free driving distance averaged over the 10 runs. Lat.\ RMSE (cross-track deviation), Term.\ err.\ (endpoint-to-goal distance), and ATE are averaged over the runs that complete the route, shown as ``--'' when none complete. RTE and RRE are computed over all 10\,m sub-sequences of all runs, including runs that end early, following the KITTI relative-error definition. The VO estimate is compared with the LiDAR ground truth.}}
\captionsetup{justification=centering, singlelinecheck=true}
\vspace{-0.6cm}
\end{table*}

\subsubsection{Dynamic Scene Robustness}
\label{subsubsec:dynamic}

Because the dense rigid BEV flow places a globally consistent rigid-motion target on every BEV cell, it is designed to steer the network toward the dominant ego-motion and to resist the locally inconsistent motion of dynamic objects. We verify this robustness on NCLT and Oxford by computing the per-frame dynamic object pixel ratio with a \rev{Mask2Former~\cite{cheng2022masked} semantic segmentation model with a Swin-L backbone, pretrained on the Cityscapes urban-scene dataset~\cite{cordts2016cityscapes}}. As shown in Fig.~\ref{fig:dynamic_error}, neither the flow EPE nor the \rev{per-frame translation error} exhibits systematic degradation as the dynamic ratio increases on either dataset, even on Oxford where dynamic content reaches 35\%.

Two structural properties of the BEV representation explain this robustness. First, the BEV projection compresses the apparent area of dynamic objects into only a few grid cells, substantially limiting the fraction of incorrectly supervised pixels. Second, the dense flow supervision covers the entire BEV grid where static cells vastly outnumber dynamic ones, so the network is driven to learn features that encode the spatially dominant ego-motion pattern.

\subsubsection{Computational Cost Analysis}

Table~\ref{table:computational_cost} reports the inference speed, GPU memory consumption, and GPU utilization for all methods under two input resolutions corresponding to NCLT (224$\times$384) and Oxford (320$\times$640). As in Tables~\ref{table:nclt_oxford_comparison}--\ref{table:kitti_zjh}, the methods are grouped by training protocol.

BEV-ODOM2 achieves 69.52\,FPS and 56.62\,FPS at the two resolutions, both well above \rev{the 16\,Hz camera rate of Oxford, the highest among the evaluated datasets}. Compared to BEV-ODOM, the addition of the PV branch introduces only a mild reduction in FPS with nearly identical GPU memory consumption, a trade-off justified by the substantial accuracy gains in Table~\ref{table:nclt_oxford_comparison}. Meanwhile, the methods based on geometric optimization or reconstruction priors, namely DROID-SLAM, DPVO, MASt3R-SLAM, and VGGT-SLAM, operate at only 7--19\,FPS, and DROID-SLAM and VGGT-SLAM consume over 14\,GB of GPU memory. BEV-ODOM2 thus achieves a favorable balance between computational efficiency and estimation accuracy, and its low memory footprint makes it well suited for resource-constrained embedded platforms.

We further validate this by deploying BEV-ODOM2 on an NVIDIA Jetson AGX Orin, where the FP32 model runs at \rev{16.91\,FPS at 224$\times$384 and 14.14\,FPS at 320$\times$640. Both rates surpass} the 10\,FPS minimum real-time constraint derived from reaction-time and stopping-distance analysis for ground robots~\cite{lin2018architectural}. Converting only the ResNet-50 backbone to TensorRT FP16 boosts the throughput to \rev{21.84 and 16.73\,FPS (29\% and 18\% speed-up)} with reduced memory, indicating substantial room for further optimization through full-model acceleration. The conversion leaves RTE and RRE unchanged to two decimal places on both datasets, so the speed-up incurs no measurable accuracy loss. \rev{With the accelerated backbone, both settings exceed the camera rates of the corresponding datasets, 5\,Hz for NCLT and 16\,Hz for Oxford. The robot in Section~\ref{subsec:closedloop} uses a 15\,Hz camera and a 320$\times$576 input. This input has fewer pixels than the 320$\times$640 setting, which already runs above 15\,FPS with the accelerated backbone.}

\renewcommand{\arraystretch}{0.95}  %
\begin{table*}[thbp]
    \centering
    \caption{ABLATION STUDY ON THREE DATASETS}
    \vspace{-5pt}
    \begin{adjustbox}{max width=0.95\textwidth}
    \begin{tabular}{
        >{\centering\arraybackslash}p{1.8cm}
        >{\centering\arraybackslash}p{1.8cm}
        >{\centering\arraybackslash}p{1.8cm}
        |>{\centering\arraybackslash}c
        >{\centering\arraybackslash}c
        >{\centering\arraybackslash}c
        |>{\centering\arraybackslash}c
        >{\centering\arraybackslash}c
        >{\centering\arraybackslash}c
        |>{\centering\arraybackslash}c
        >{\centering\arraybackslash}c
        >{\centering\arraybackslash}c
    }
    \toprule
    \multirow{3}{*}{\parbox{1.8cm}{\centering \textbf{Enhanced \\ Rotation \\ Sampling}}} & 
    \multirow{3}{*}{\parbox{1.8cm}{\centering \textbf{Dense Flow \\ Supervision}}} & 
    \multirow{3}{*}{\parbox{1.8cm}{\centering \textbf{PV-BEV \\ Fusion}}} & 
    \multicolumn{3}{c|}{\textbf{NCLT}} & 
    \multicolumn{3}{c|}{\textbf{Oxford}} & 
    \multicolumn{3}{c}{\textbf{ZJH-VO}} \\
    \noalign{\vskip 0.5mm}
    \cline{4-6} \cline{7-9} \cline{10-12}
    \noalign{\vskip 0.75mm}
    & & & 
    \textbf{RTE*} & 
    \textbf{RRE*} & 
    \textbf{ATE$^{\dag}$} & 
    \textbf{RTE*} & 
    \textbf{RRE*} & 
    \textbf{ATE$^{\dag}$} & 
    \textbf{RTE*} & 
    \textbf{RRE*} & 
    \textbf{ATE$^{\dag}$} \\
    & & & 
    \textbf{(\%)} & 
    \textbf{(°/100\,m)} & 
    \textbf{(m)} & 
    \textbf{(\%)} & 
    \textbf{(°/100\,m)} & 
    \textbf{(m)} & 
    \textbf{(\%)} & 
    \textbf{(°/100\,m)} & 
    \textbf{(m)} \\
    \midrule
    \ding{55} & \ding{55} & \ding{55} & 9.95 & 4.31 & 126.63 & 8.67 & 1.74 & 108.35 & 7.28 & 7.98 & 3.68 \\
    \ding{51} & \ding{55} & \ding{55} & \rev{7.31} & \rev{3.28} & \rev{128.68} & \rev{6.71} & 1.17 & \rev{108.75} & \rev{6.29} & \rev{7.40} & \rev{2.57} \\
    \ding{51} & \ding{51} & \ding{55} & \rev{5.17} & \rev{2.25} & \rev{75.77} & \rev{5.56} & \rev{1.10} & \rev{\textbf{66.71}} & \rev{4.02} & \rev{6.14} & \rev{1.86} \\
    \ding{51} & \ding{55} & \ding{51} & \rev{5.83} & \rev{2.73} & \rev{145.94} & \rev{5.37} & \rev{1.12} & \rev{84.96} & \rev{5.12} & \rev{5.64} & \rev{\textbf{1.83}} \\
    \ding{51} & \ding{51} & \ding{51} & \rev{\textbf{4.68}} & \rev{\textbf{2.19}} & \rev{\textbf{65.75}} & \rev{\textbf{4.84}} & \rev{\textbf{1.00}} & \rev{71.34} & \rev{\textbf{3.65}} & \rev{\textbf{5.58}} & \rev{1.89} \\
    \bottomrule
    \end{tabular}
    \end{adjustbox}
    \label{table:ablation_study}
    \captionsetup{justification=raggedright, singlelinecheck=false}
    \caption*{\footnotesize{\textbf{Enhanced Rotation Sampling}: Rotation-aware sampling strategy to balance diverse motion patterns and address dataset bias. \\
    \textbf{Dense Flow Supervision}: \rev{Dense rigid BEV flow computed from the pose ground truth, providing a pixel-level training target.} \\
    \textbf{PV-BEV Fusion}: Dual-branch fusion that \rev{retains additional motion cues from the PV correlation computed before projection.} \\
    $^{*}$ Aligned using SE(3); $^{\dag}$ aligned using Sim(3).}}
    \captionsetup{justification=centering, singlelinecheck=true}
\vspace{-0.6cm}
\end{table*}
\renewcommand{\arraystretch}{1.0}

\subsection{Closed-Loop Navigation Validation}
\label{subsec:closedloop}

The preceding sections evaluate odometry accuracy offline. This section examines whether this accuracy holds in a real navigation task. Both BEV-ODOM2 and the BEV-ODOM baseline are deployed as the sole localization source of a teach-and-repeat system on a physical ground robot. The platform is the same four-wheeled vehicle that collected ZJH-VO (Appendix~\ref{appendix:zjhvo}), and a single forward-facing monocular camera provides the only input, streaming $1280\times800$ images at $15$\,Hz. The global pose is integrated online from frame-to-keyframe motion, with the keyframe refreshed every $0.5$\,s to reduce error accumulation.

\begin{figure}[t]
\centering
\includegraphics[width=0.98\columnwidth]{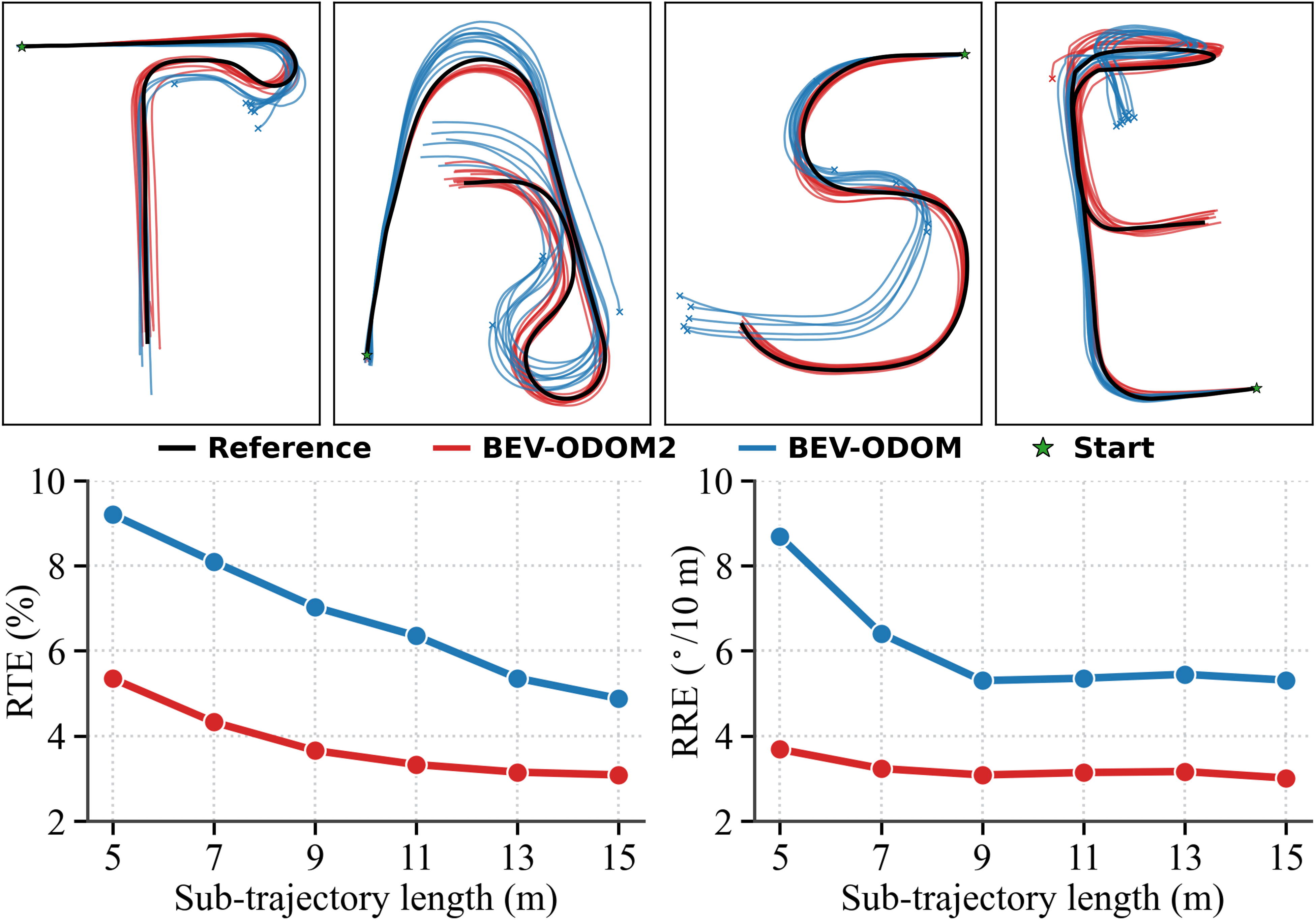}
\caption{Closed-loop route following on routes T, A, S, and E. Top: taught reference (black) and the driven trajectories (LiDAR ground truth) of BEV-ODOM2 (red) and BEV-ODOM (blue) over the ten repeats, all drawn in the shared LiDAR map frame without any trajectory alignment; the green star marks each start. Bottom: RTE (\%) and RRE (°/10\,m) of the VO estimate against the LiDAR ground truth as a function of the sub-trajectory length, aggregated over all runs and routes.}
\label{fig:closedloop}
\vspace{-0.6cm}
\end{figure}

We first build a map of the site with LiDAR SLAM. During teaching, LiDAR localization runs on this map while we drive the robot manually along a chosen route, and the localized poses are recorded at a $0.1$\,m spacing to form the reference path. During repetition, the robot starts near the beginning of the route and the controller is driven entirely by the visual odometry. This setting emulates a localization dropout: the odometry is initialized once from the current LiDAR pose when it takes over, and receives no LiDAR correction thereafter. The reference path and the driven trajectory therefore share the map frame, and the gap between them directly reflects the accumulated odometry drift. The trajectories in Fig.~\ref{fig:closedloop} carry no post-hoc alignment. Throughout the repetition, the LiDAR only logs the ground truth and never enters the control loop.

Both models are initialized from their ZJH-VO weights and fine-tuned on ten minutes of new-scene data, about $0.5$ GPU-hours in total. The fine-tuning data covers less than $300$\,m of trajectory, a small fraction of the more than $3{,}000$\,m in ZJH-VO. We evaluate on four routes labeled T, A, S, and E, with lengths from $12.9$ to $23.1$\,m as listed in Table~\ref{table:closedloop_success}. Each route is taught once and then repeated $10$ times per method under a pure-pursuit controller, for $80$ autonomous runs. A run ends at the first collision or human intervention, and a run that reaches the end of the route with neither is counted as complete.

Table~\ref{table:closedloop_success} reports two metric groups: the closed-loop task error of the driven path against the taught reference, and the odometry error of the VO estimate against the LiDAR ground truth. The task error combines the odometry drift with the control tracking error.

BEV-ODOM2 completes 39 of the 40 runs and drives every route with a cross-track RMSE of $7$--$15$\,cm. It stays collision-free over the full length of routes T, A, and S, and over $22.4$\,m of the $23.1$\,m route E on average, while BEV-ODOM accumulates a larger drift that often pushes it off the path before the route ends. The top panels of Fig.~\ref{fig:closedloop} show this directly: the trajectories of BEV-ODOM2 stay close to each taught path in the shared map frame, while the baseline leaves the route, most visibly on routes S and E. The bottom panels report RTE and RRE against the sub-trajectory length, where BEV-ODOM2 stays well below BEV-ODOM at every length on both metrics. Even on the runs where the baseline completes the route, its ATE and RTE remain well above ours. In Table~\ref{table:closedloop_success}, the only metric where the baseline is better is the RRE on route T. There, the baseline runs mostly end in collision soon after the initial long straight, so the measurable segments are dominated by straight driving with little rotation. Overall, these results show that the offline accuracy gain of BEV-ODOM2 carries over into higher task reliability.

\subsection{Ablation Study}
\label{subsec:ablation}

To verify the effectiveness of each core component in BEV-ODOM2, we conduct ablation experiments on NCLT, Oxford, and our collected ZJH-VO dataset, with results shown in Table~\ref{table:ablation_study}. \rev{Row 1 contains none of the three components and corresponds to BEV-ODOM \cite{wei2024bev}. Starting from this baseline,} we add the enhanced rotation sampling (row 2) and then the dense flow supervision (row 3). Row 4 applies the PV-BEV fusion on top of the sampling strategy alone, isolating the fusion from the dense flow supervision, and row 5 is the full model.

\textbf{Effect of Enhanced Rotation Sampling:} Introducing the enhanced rotation sampling strategy (row 2) \rev{improves RTE and RRE on all datasets.} This confirms that balancing the motion patterns in the training data effectively mitigates dataset bias and enhances the model's ability to estimate less frequent but critical actions like turns.

\textbf{Effect of Dense Flow Supervision:} Adding the dense rigid BEV flow supervision (row 3) \rev{reduces RTE by 29.3\%, 17.1\%, and 36.1\% on NCLT, Oxford, and ZJH-VO, respectively.} Unlike sparse pose supervision that constrains only a single motion vector per frame, the dense flow places a learning target on every BEV cell, providing much richer gradient signals during training. \rev{This row adds the flow decoder together with its target. Predicting a pixel-level correspondence at every cell} strengthens the network's learning of fine motion patterns, especially in scenes spanning a wide range of object distances. The consistent ATE gains across all three datasets further indicate that predicted trajectories stay closer to ground truth, confirming more stable pose estimation.

\textbf{Effect of PV-BEV Fusion:} Adding the PV branch on top of sampling alone (row 4 versus row 2) \rev{reduces RTE by 20.2\%, 20.0\%, and 18.6\% on NCLT, Oxford, and ZJH-VO, respectively. It also lowers RRE on all three datasets, so the PV correlation is effective} even without the dense flow supervision. This gain reflects an overall enrichment of the learned motion representation during training rather than a per-frame correction of high-pitch or high-roll motions.

\textbf{Full Model:} Combining the dense flow supervision and the PV-BEV fusion (row 5) \rev{gives the lowest RTE and RRE on all three datasets.}

\section{Conclusion and Future Work}
This paper introduces BEV-ODOM2, a monocular visual odometry framework that improves upon the earlier methods in the BEV-ODOM line through two key components: pose-derived dense rigid BEV flow supervision that turns pose ground truth into pixel-level guidance, and a PV-BEV fusion strategy that \rev{computes correlation before projection to retain additional motion cues and help alleviate projection ambiguity}. An enhanced rotation sampling strategy further ensures robust performance across diverse motion patterns.

We evaluate BEV-ODOM2 on four datasets that span diverse outdoor and indoor environments with varied spatial scales. KITTI and Oxford cover suburban and urban road driving, NCLT represents campus-level navigation, and our newly collected ZJH-VO dataset covers smaller-scale structured environments, including an underground garage, an outdoor plaza, corridors, and dense offices. \rev{On these datasets, BEV-ODOM2 reduces the relative translation error of BEV-ODOM by 39\% on average, with reductions of 53.0\%, 44.2\%, 9.4\%, and 49.9\% on NCLT, Oxford, KITTI, and ZJH-VO. It also maintains} the robust scale consistency inherent to the BEV representation. In closed-loop navigation on a physical robot, BEV-ODOM2 serves as the sole localization source and follows all four taught routes with a cross-track RMSE within $7$--$15$\,cm, showing that the accuracy gain carries over to task-level reliability. A dead-reckoning analysis in Appendix~\ref{appendix:dr} further \rev{shows that BEV-ODOM2 achieves the highest \rev{mean} availability under the lateral tolerance of each scenario during 10-second positioning outages. This indicates} its suitability as a localization bridge during such intervals.

To facilitate future research, we open-source both the BEV-ODOM2 code and the ZJH-VO dataset, together with calibration data and performance baselines from representative MVO methods.

Future work will focus on enhancing long-term robustness by integrating multi-camera systems for 360-degree perception, incorporating relocalization and loop closure capabilities to correct accumulated drift, and developing adaptive temporal models for stability in adverse conditions. In addition, although BEV-ODOM2 already runs above 20\,FPS \rev{with 224$\times$384 input} on a Jetson AGX Orin with partial TensorRT acceleration, full model quantization and network pruning will be explored to further reduce latency for real-time operation on resource-constrained embedded platforms. These directions aim to transition BEV-ODOM2 from a research prototype to a production-ready navigation module for autonomous ground robots operating across diverse environments.

\bibliographystyle{IEEEtran}
\bibliography{references}

\clearpage
\appendices

\section{ZJH-VO Dataset Details}
\label{appendix:zjhvo}

This appendix provides detailed descriptions of the ZJH-VO dataset, including the data collection platform, environment characteristics, and motion statistics. The main text (Section~\ref{subsec:datasets}) summarizes the dataset design rationale and training/testing split; this appendix supplements the full specification for reproducibility.

\begin{figure*}[t]
\centering
\includegraphics[width=0.98\textwidth]{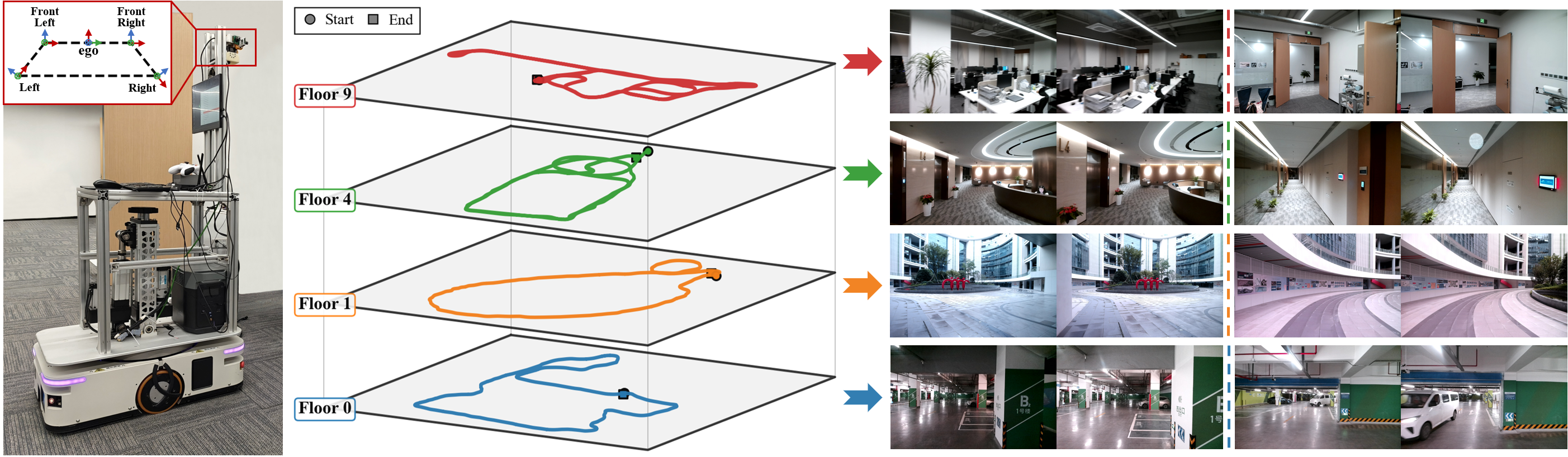}
\caption{ZJH-VO Multi-Scale Dataset Overview. Left: Data collection platform and camera coordinate system setup with PV-BEV transformation. Center: Representative trajectory paths across four floors (Floor 9, 4, 1, and 0). Right: Sample image pairs from each environment showing diverse scene characteristics.}
\label{fig:ZJH-VO_dataset_intro}
\vspace{-0.4cm}
\end{figure*}

The dataset is recorded across multiple floors of an office complex using a four-wheeled mobile platform without suspension, which amplifies platform jitter and increases the difficulty of visual odometry. The platform carries a hardware-synchronized four-camera system and a 2D LiDAR for ground truth generation. The camera system consists of a front-facing stereo pair and two additional front-left and front-right cameras with approximately 30-degree overlapping fields of view. We use only the left camera of the front stereo pair in our experiments.

\textbf{Calibration and synchronization.} Camera intrinsics are calibrated with a checkerboard target and reprojection-error screening. The camera-to-LiDAR extrinsics are calibrated by aligning geometric features observed by both sensors and refined by minimizing the alignment residuals. All cameras and the LiDAR are triggered by a shared hardware clock, so image and scan timestamps come from the same time base and no software-level time alignment is involved.

\textbf{Ground truth generation and accuracy.} Ground-truth poses are obtained by localizing the 2D LiDAR against pre-built occupancy maps of each floor. The maps are constructed offline with loop closures, and the localization runs scan matching against these fixed maps, so the ground truth does not drift with trajectory length. The map-based localization is also independent of the evaluated odometry, and all sequences on the same floor share one fixed map, so the reference is consistent across sequences. This localization is the factory navigation system of our commercial AGV base, rated by the vendor at a repeat positioning accuracy of $\pm10$\,mm and $\pm1^{\circ}$.

\textbf{Validity of the planar ground truth.} The 2D LiDAR ground truth provides 3-DoF poses. This form fits the platform: the vehicle is a rigid four-wheeled platform without suspension, and every recorded floor is flat, so the out-of-plane motion is limited to small jitter without sustained slope. The 3-DoF ground truth therefore closely reflects the true motion.

\begin{figure}[t]
\centering
\includegraphics[width=0.5\textwidth]{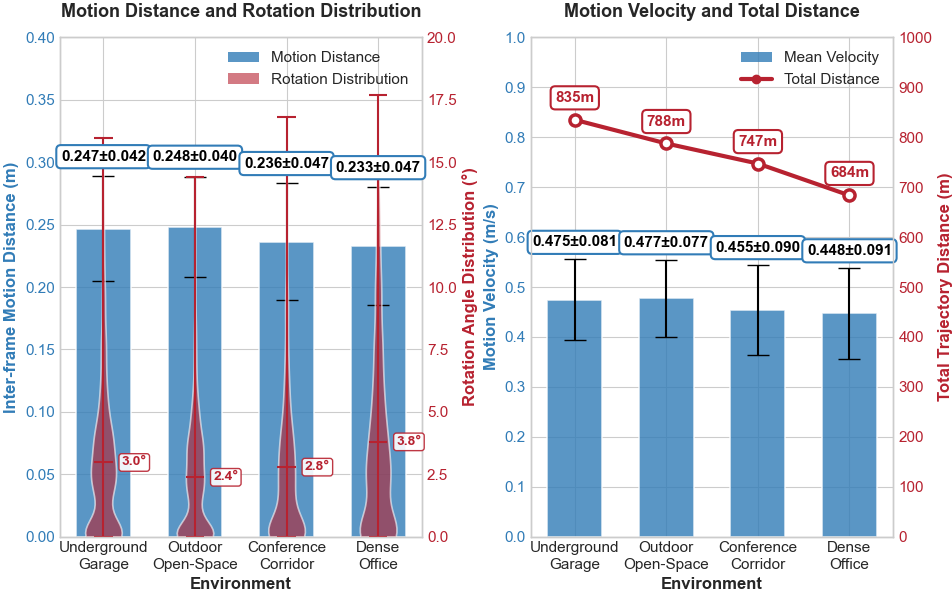}
\caption{ZJH-VO Multi-Scale dataset motion statistics across four environments. Left: Inter-frame motion distance and rotation angle distributions with statistical measures. Right: Mean motion velocity and total trajectory distance for each environment sequence.}
\label{fig:ZJH-VO_analysis}
\vspace{-0.4cm}
\end{figure}

The dataset covers four representative environments with increasing spatial constraints: an underground garage, an outdoor plaza, corridors, and dense offices. Each environment contains 3 trajectories with unique coverage areas, totaling 12 trajectories \rev{and 3,054 meters of motion data. The images are stored at the full camera rate of 15\,Hz. \rev{Training pairs are drawn from these images by the enhanced rotation sampling of Section~\ref{subsec:enhanced_strategy}}, and evaluation uses every eighth frame. At this interval, the 12 trajectories contain 12,666 frames.} \rev{Two further trajectories, recorded separately, are released for validation.} Fig.~\ref{fig:ZJH-VO_dataset_intro} illustrates the data collection platform, trajectory examples from different floors, and sample frames as examples.

The four environments pose different challenges to BEV-based odometry. The underground garage has moving vehicles and dim lighting. The outdoor plaza is open and dominated by ground cues, and most landmarks are more than 10\,m away. The corridors are narrow, with variable lighting and floor-to-ceiling glass windows. The dense offices are cluttered, with furniture within 0.5\,m to 3\,m of the camera, textureless walls, and frequent large in-place rotations. A single model must generalize across all four environments, which span a wide range of observation distances and motion patterns. This motivates the multi-scene design of ZJH-VO.

Fig.~\ref{fig:ZJH-VO_analysis} presents motion characteristics across four environments through dual-axis visualization. The left panel shows inter-frame motion distances (bars) and rotation angle distributions (violin plots), while the right panel displays motion velocities (bars) and total trajectory distances (line plot). Across all environments, the distances between evaluated frames range from 0.00 to 0.31\,m, their rotation angles from 0.00° to 17.70°, and the motion velocities from 0.00 to 0.59\,m/s.

\section{Hyperparameter \rev{Sensitivity}}
\label{appendix:hyperparam}

\rev{\textbf{Loss Weights:}} Table~\ref{table:hyperparameter_sensitivity}(a) \rev{reports the sensitivity to} $\lambda_1$ and $\lambda_2$, with $\alpha{=}10$ and $\beta{=}10$ fixed. These factors balance the rotation and translation terms within each loss. Both auxiliary losses are designed to provide supplementary guidance rather than dominate training: increasing either weight ($\lambda_1{:}\,0.2{\to}1.0$ or $\lambda_2{:}\,0.1{\to}0.5$) degrades performance, as overly strong auxiliary supervision diverts the shared feature representation away from the primary 3-DoF BEV pose objective.

\rev{\textbf{Sampling Ratio:}} Table~\ref{table:hyperparameter_sensitivity}(b) compares three sampling distributions. Equal sampling (50:50) leaves the model underexposed to turning maneuvers, while the rotation-dominant ratio (90:10) starves straight-line training, which harms generalization because straight segments dominate real-world driving. The adopted 70:30 ratio \rev{gives the lowest errors among the three settings}. This experiment is conducted on Oxford, whose predominantly straight road segments make it particularly diagnostic for evaluating sampling ratio effects.

\renewcommand{\arraystretch}{0.95}
\begin{table}[t]
    \centering
    \caption{\rev{HYPERPARAMETER SENSITIVITY ON OXFORD DATASET}}
    \vspace{-5pt}
    \begin{adjustbox}{max width=\columnwidth}
    \begin{tabular}{
        >{\centering\arraybackslash}p{0.55cm}
        >{\centering\arraybackslash}p{0.75cm}|
        >{\centering\arraybackslash}p{0.65cm}
        >{\centering\arraybackslash}p{1cm}|
        >{\centering\arraybackslash}p{1.3cm}|
        >{\centering\arraybackslash}p{0.65cm}
        >{\centering\arraybackslash}p{1cm}
    }
    \toprule
    \multicolumn{4}{c|}{\textbf{(a) Loss Weight}} & \multicolumn{3}{c}{\textbf{(b) Sampling Ratio}} \\
    \midrule
    $\lambda_1$ & $\lambda_2$ & \textbf{RTE*} & \textbf{RRE*} & \textbf{High:Std} & \textbf{RTE*} & \textbf{RRE*} \\
     (PV) &  (Flow) & (\%) & (°/100\,m) & Ratio & (\%) & (°/100\,m) \\
    \midrule
    1.0 & 0.1 & 5.18 & 1.26 & 50:50 & 5.28 & 1.18 \\
    0.2 & 0.5 & 5.34 & 1.10 & \textbf{70:30} & \rev{\textbf{4.84}} & \rev{\textbf{1.00}} \\
    \textbf{0.2} & \textbf{0.1} & \rev{\textbf{4.84}} & \rev{\textbf{1.00}} & 90:10 & 5.45 & 1.12 \\
    \bottomrule
    \end{tabular}
    \end{adjustbox}
    \label{table:hyperparameter_sensitivity}
    \captionsetup{justification=raggedright, singlelinecheck=false}
    \caption*{\footnotesize{All variants use the full BEV-ODOM2 architecture. Results are averaged over four Oxford test sequences. Bold denotes the adopted configuration. \\
    $^{*}$ Aligned using SE(3).}}
    \captionsetup{justification=centering, singlelinecheck=true}
\vspace{-0.8cm}
\end{table}
\renewcommand{\arraystretch}{1.0}

\renewcommand{\arraystretch}{1.0}
\begin{table*}[t]
    \centering
    \caption{LATERAL (CROSS-TRACK) DEAD-RECKONING ERROR @ 10\,s (UNIT: m)}
    \vspace{-5pt}
    \resizebox{\textwidth}{!}{
    \begin{tabular}{c|cc|cccc|cccc|cccc}
    \toprule
    \multirow{3}{*}[-5pt]{\textbf{Method}} & \multicolumn{6}{c|}{\textbf{Tier 1: Suburban \& Urban Driving}} & \multicolumn{4}{c|}{\textbf{Tier 2: Campus-Level Navigation}} & \multicolumn{4}{c}{\textbf{Tier 3: Small-Scale Env.}} \\
    \cmidrule(lr){2-7} \cmidrule(lr){8-11} \cmidrule(lr){12-15}
    & \multicolumn{2}{c|}{\textbf{KITTI}} & \multicolumn{4}{c|}{\textbf{Oxford}} & \multicolumn{4}{c|}{\textbf{NCLT}} & \multicolumn{4}{c}{\textbf{ZJH-VO}} \\
    \cmidrule(lr){2-3} \cmidrule(lr){4-7} \cmidrule(lr){8-11} \cmidrule(lr){12-15}
      & \textbf{Seq09} & \textbf{Seq10}
      & \textbf{11-12} & \textbf{15-13} & \textbf{16-14} & \textbf{17-12}
      & \textbf{02-02} & \textbf{02-19} & \textbf{03-17} & \textbf{08-20}
      & \textbf{F-09} & \textbf{F-04} & \textbf{F-01} & \textbf{F-00} \\
    \midrule
    BEV-DWPVO~\cite{wei2025bev}
      & 2.03 & 1.45
      & \underline{0.39} & \underline{0.41} & \underline{0.36} & \underline{0.50}
      & 0.22 & 0.20 & 0.20 & \underline{0.21}
      & 0.08 & 0.07 & 0.09 & 0.07 \\
    BEV-ODOM~\cite{wei2024bev}
      & \textbf{0.83} & \underline{1.04}
      & 0.58 & 0.46 & 0.61 & 0.53
      & \underline{0.20} & \underline{0.16} & \underline{0.17} & \underline{0.21}
      & \underline{0.05} & \textbf{0.04} & \underline{0.04} & \underline{0.06} \\
    \textbf{BEV-ODOM2 (Ours)}
      & \underline{1.01} & \textbf{0.89}
      & \textbf{0.36} & \textbf{0.33} & \textbf{0.33} & \textbf{0.37}
      & \textbf{0.17} & \textbf{0.13} & \textbf{0.14} & \textbf{0.16}
      & \textbf{0.04} & \textbf{0.04} & \textbf{0.03} & \textbf{0.05} \\
    \bottomrule
    \end{tabular}
    }
    \label{table:lateral_dr_error}
    \captionsetup{justification=raggedright, singlelinecheck=false}
    \caption*{\footnotesize{Median lateral error accumulated over 10\,s of dead-reckoning. The 10\,s window represents a typical duration for brief GNSS outages encountered in tunnels, parking structures, and building interiors~\cite{li2017tightly}.}}
    \captionsetup{justification=centering, singlelinecheck=true}
\vspace{-0.4cm}
\end{table*}
\renewcommand{\arraystretch}{1.0}

\renewcommand{\arraystretch}{1.0}
\begin{table*}[t]
    \centering
    \caption{LATERAL LOCALIZATION AVAILABILITY @ 10\,s (\%)}
    \vspace{-5pt}
    \resizebox{\textwidth}{!}{
    \begin{tabular}{c|cc|cccc|cccc|cccc}
    \toprule
    \multirow{3}{*}[-5pt]{\textbf{Method}} & \multicolumn{6}{c|}{\textbf{Tier 1: Suburban \& Urban Driving}} & \multicolumn{4}{c|}{\textbf{Tier 2: Campus-Level Navigation}} & \multicolumn{4}{c}{\textbf{Tier 3: Small-Scale Env.}} \\
    \cmidrule(lr){2-7} \cmidrule(lr){8-11} \cmidrule(lr){12-15}
    & \multicolumn{2}{c|}{\textbf{KITTI ($<$3.66\,m)}} & \multicolumn{4}{c|}{\textbf{Oxford ($<$1.83\,m)}} & \multicolumn{4}{c|}{\textbf{NCLT ($<$0.50\,m)}} & \multicolumn{4}{c}{\textbf{ZJH-VO ($<$0.25\,m)}} \\
    \cmidrule(lr){2-3} \cmidrule(lr){4-7} \cmidrule(lr){8-11} \cmidrule(lr){12-15}
      & \textbf{Seq09} & \textbf{Seq10}
      & \textbf{11-12} & \textbf{15-13} & \textbf{16-14} & \textbf{17-12}
      & \textbf{02-02} & \textbf{02-19} & \textbf{03-17} & \textbf{08-20}
      & \textbf{F-09} & \textbf{F-04} & \textbf{F-01} & \textbf{F-00} \\
    \midrule
    BEV-DWPVO~\cite{wei2025bev}
      & 83.4 & \underline{95.9}
      & \underline{95.4} & \underline{96.8} & \textbf{97.7} & \underline{90.4}
      & 80.4 & 83.1 & 82.2 & 81.5
      & 80.7 & 85.1 & 75.7 & 88.2 \\
    BEV-ODOM~\cite{wei2024bev}
      & \underline{98.8} & 92.7
      & 88.0 & 92.1 & 87.6 & 90.0
      & \underline{85.9} & \underline{87.3} & \underline{86.4} & \underline{84.5}
      & \underline{92.8} & \underline{91.7} & \underline{98.2} & \underline{92.1} \\
    \textbf{BEV-ODOM2 (Ours)}
      & \textbf{100.0} & \textbf{97.0}
      & \textbf{95.9} & \textbf{98.0} & \underline{97.3} & \textbf{94.8}
      & \textbf{90.4} & \textbf{94.2} & \textbf{92.9} & \textbf{94.2}
      & \textbf{96.3} & \textbf{94.3} & \textbf{98.7} & \textbf{97.7} \\
    \bottomrule
    \end{tabular}
    }
    \label{table:localization_availability}
    \captionsetup{justification=raggedright, singlelinecheck=false}
    \caption*{\footnotesize{Fraction of 10\,s DR sub-trajectories with lateral error below the tier-specific ceiling. Ceilings: 3.66\,m = one standard lane width for suburban roads~\cite{reid2019localization}; 1.83\,m = half a lane width for urban roads; 0.50\,m = campus path half-width~\cite{rehrl2021evaluating, iso22737}; 0.25\,m = valet-parking lateral tolerance~\cite{iso23374, wenzel20254seasons}.}}
    \captionsetup{justification=centering, singlelinecheck=true}
\vspace{-0.4cm}
\end{table*}
\renewcommand{\arraystretch}{1.0}

\section{Dead-Reckoning Application Analysis}
\label{appendix:dr}

GNSS signals are frequently unavailable in tunnels, parking structures, building interiors, and urban canyons. During such outages, the navigation system must fall back to dead-reckoning (DR) to maintain localization~\cite{reid2019localization}. The value of a better visual odometry module is therefore reflected by how accurately it can maintain the platform's estimated position during these GNSS-denied intervals. Building on the standard trajectory-level ATE/RTE metrics in Section~\ref{subsec:experimental_setup}, we introduce two complementary DR-oriented metrics that directly characterize this capability.

\textbf{Lateral Dead-Reckoning Error.}
We decompose the position error accumulated over a fixed time window into lateral (cross-track) and longitudinal (along-track) components relative to the ground-truth heading. The \emph{lateral} component determines whether the platform drifts out of its traversable path, which is the \rev{main concern for path keeping} during a positioning outage~\cite{reid2019localization}. We adopt a 10-second evaluation window, representative of brief outages caused by overpasses, short tunnels, and indoor transitions~\cite{li2017tightly}. For each sequence, we sample up to 500 starting points and report the median lateral error across all sub-trajectories.

\textbf{Localization Availability.}
We define localization availability as the fraction of DR sub-trajectories whose \emph{lateral} position error remains below a scenario-specific ceiling. This metric, adapted from the integrity framework of Reid et al.~\cite{reid2019localization} and the availability-based evaluation of Rehrl and Gr\"{o}chenig~\cite{rehrl2021evaluating}, provides a single percentage summarizing how often the DR bridge keeps the platform within its traversable path. We set tier-specific lateral ceilings based on the physical constraints of each scenario: for Tier~1, we use 3.66\,m on KITTI (one standard lane width on suburban roads~\cite{reid2019localization}) and 1.83\,m on Oxford (half a lane width on urban roads with tighter clearances); 0.50\,m for Tier~2 campus-level navigation (half the typical path width, aligned with the lateral boundary constraints of low-speed automated driving systems~\cite{iso22737, rehrl2021evaluating}); and 0.25\,m for Tier~3 small-scale structured environments, which encompass an underground garage, an outdoor plaza, corridors, and dense offices. This ceiling is consistent with the 0.20--0.30\,m lateral tolerance commonly adopted for automated valet parking~\cite{iso23374} and falls within the medium-accuracy tier of the 4Seasons benchmark~\cite{wenzel20254seasons}. \rev{These ceilings are application-motivated reference tolerances for evaluation, not formal safety guarantees.}

Tables~\ref{table:lateral_dr_error} and \ref{table:localization_availability} report the lateral DR error and the corresponding localization availability at 10\,s for the three BEV-based methods across four datasets. We analyze each tier below, from the widest to the strictest lateral ceiling.

\textbf{Tier~1: Suburban and Urban Driving (KITTI, Oxford).}
On KITTI, BEV-ODOM2 yields median lateral errors of 1.01\,m and 0.89\,m on Seq09 and Seq10, both well within the 3.66\,m full-lane ceiling. On Oxford, the median lateral error across all four sequences is 0.33--0.37\,m, about one-fifth of the 1.83\,m half-lane ceiling. The corresponding availability reaches 97.0--100.0\% on KITTI and 94.8--98.0\% on Oxford; across all six Tier~1 sequences the availability exceeds 94\%, a 5.6 percentage-point gain over BEV-ODOM and a 3.9 percentage-point gain over BEV-DWPVO in mean availability. These results confirm that BEV-ODOM2 keeps the estimated position within its path during a 10\,s positioning outage on nearly all evaluated suburban and urban road segments.

\textbf{Tier~2: Campus-Level Navigation (NCLT).}
Under the 0.50\,m campus-path ceiling, BEV-ODOM2 achieves median lateral errors of 0.13--0.17\,m across all four NCLT sequences, 15--35\% lower than both baselines. The mean availability reaches 92.9\%, a 6.9 percentage-point gain over BEV-ODOM and an 11.1 percentage-point gain over BEV-DWPVO. All four median errors remain below the 0.50\,m lateral ceiling, which aligns with the low-speed automated driving boundary constraints in ISO~22737~\cite{iso22737}. The availability figures meet the empirical positioning benchmarks reported by Rehrl and Gr\"{o}chenig~\cite{rehrl2021evaluating}, confirming that BEV-ODOM2 can reliably bridge short positioning outages in campus environments.

\textbf{Tier~3: Small-Scale Structured Environments (ZJH-VO).}
In the most constrained environments, BEV-ODOM2 achieves median lateral errors of 0.03--0.05\,m, well below the 0.25\,m ceiling reflecting valet-parking-level lateral requirements~\cite{iso23374}. The mean availability reaches 96.8\%, a 14.3 percentage-point gain over BEV-DWPVO and a 3.1 percentage-point gain over BEV-ODOM. These \rev{median} errors of at most 5\,cm keep the estimated position within corridors and passageways, supporting navigation in GNSS-denied building interiors.

\begin{figure*}[t]
\vspace{-3pt}
\centering
\includegraphics[width=0.90\textwidth]{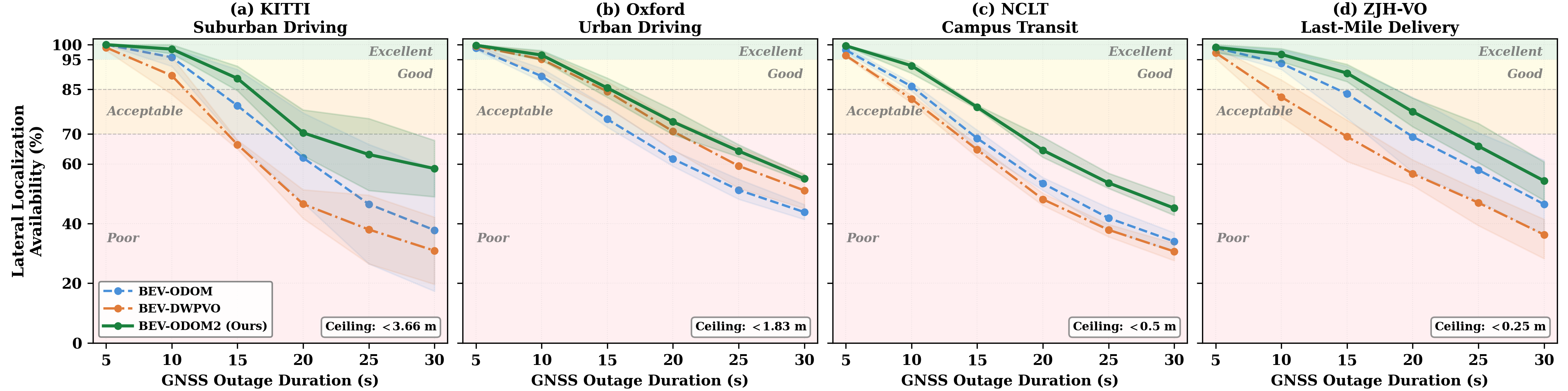}
\vspace{-0.2cm}
\caption{Lateral localization availability degradation over GNSS outage duration (5--30\,s) on four datasets. Each panel shows the mean availability (solid line) and sequence-level min--max range (shaded band) for three BEV-based methods. Background bands indicate quality tiers: Excellent ($\geq$95\%), Good (85--95\%), Acceptable (70--85\%), and Poor ($<$70\%). Lateral ceilings are set per scenario: 3.66\,m for KITTI suburban driving, 1.83\,m for Oxford urban driving, 0.50\,m for NCLT campus-level navigation, and 0.25\,m for ZJH-VO small-scale structured environments.}
\label{fig:availability_degradation}
\vspace{-0.5cm}
\end{figure*}

\textbf{Availability Degradation over Extended Outages.}
The above results are evaluated at a fixed 10\,s window. Fig.~\ref{fig:availability_degradation} further examines how availability degrades as the positioning outage extends from 5\,s to 30\,s~\cite{li2017tightly}. Across all four datasets, BEV-ODOM2 degrades more slowly than both baselines. On KITTI, BEV-ODOM2 remains in the Good range at 15\,s while BEV-DWPVO has already entered the Poor range. On Oxford, BEV-ODOM2 similarly leads both baselines throughout the 5--30\,s range. By 30\,s, all methods enter the Poor range on Tier~1 datasets, yet BEV-ODOM2 retains a 21--28 percentage-point lead on KITTI and a 4--11 percentage-point lead on Oxford. On NCLT, BEV-ODOM2 decays more slowly than the two baselines, retaining an 11--15 percentage-point advantage at 30\,s. On ZJH-VO, the stringent 0.25\,m ceiling causes all methods to descend more rapidly, yet BEV-ODOM2 consistently stays above both baselines throughout the entire range. The shaded min--max bands further show that BEV-ODOM2 exhibits narrower inter-sequence variability, indicating more uniform performance across different routes and conditions.

\textbf{Discussion.}
Across the three tiers, BEV-ODOM2 achieves the best lateral accuracy on 13 of 14 sequences and the highest mean availability on all four datasets. The advantage holds not only at 10\,s but also over extended outages up to 30\,s. These results \rev{indicate} that BEV-based odometry can serve as a viable \emph{dead-reckoning bridge} during GNSS-denied intervals. It is not a standalone positioning solution, but a module that extends \rev{the time the lateral error stays within the application tolerance} before absolute correction becomes necessary. The BEV grid's metric scale constrains the scale drift that dominates monocular errors, and the dense flow supervision together with PV-BEV fusion further \rev{reduces lateral drift}, yielding the consistent improvements of BEV-ODOM2 over BEV-ODOM and BEV-DWPVO. Table~\ref{table:computational_cost} further shows that BEV-ODOM2 runs at 21.84\,FPS \rev{with 224$\times$384 input} on a Jetson AGX Orin with \rev{a} TensorRT backbone, \rev{above the 10\,FPS minimum for ground robots}. Taken together, these results indicate that BEV-ODOM2 is suitable for deployment \rev{in the road, campus, and indoor scenarios represented by the four datasets}.

\end{document}